\documentclass{article}

\PassOptionsToPackage{numbers, compress}{natbib}

 \usepackage[preprint]{neurips_2026}

\usepackage[utf8]{inputenc} 
\usepackage[T1]{fontenc}    
\usepackage{hyperref}       
\usepackage{url}            
\usepackage{booktabs}       
\usepackage{amsfonts}       
\usepackage{nicefrac}       
\usepackage{microtype}      
\usepackage{xcolor}         
\definecolor{viridisgreen}{HTML}{5EC962}

\usepackage{pifont}
\newcommand{\cmark}{\ding{51}}%
\newcommand{\xmark}{\ding{55}}%
\usepackage{amsmath}
\usepackage{booktabs}
\usepackage{multirow}
\usepackage{makecell}
\usepackage{pifont}
\usepackage{graphicx}
\usepackage{wrapfig}
\usepackage[table]{xcolor}
\definecolor{customblue}{HTML}{4C78A8}

\title{AnyBokeh: Physics-Guided Any-to-Any Bokeh Editing with Optical Fingerprint Transfer}

\author{%
  Xinyu Hou \quad Xiaoming Li \quad Zongsheng Yue \quad Chen Change Loy \\
  S-Lab, College of Computing \& Data Science \\
  Nanyang Technological University \\
  Singapore, 639798 \\
  \texttt{\{xinyu.hou, ccloy\}@ntu.edu.sg}, \texttt{\{csxmli, zsyzam\}@gmail.com}
}

\begin{document}

\maketitle

\begin{abstract}
    Depth-of-field control is a fundamental tool in photography, yet post-capture bokeh editing from a single image remains challenging. A practical editor should handle images captured under arbitrary focus and aperture settings. Existing methods typically assume an all-in-focus input, or first recover an all-in-focus image before rendering new bokeh. Such pipelines can discard useful blur cues from the source image and propagate reconstruction artifacts into the final edit.
    We introduce \textbf{AnyBokeh}, a physics-guided framework for \textbf{any-to-any bokeh editing}. Instead of treating source blur merely as a degradation to be removed, AnyBokeh estimates the \textbf{source blur state} with a signed circle-of-confusion map and a disparity map. By modeling the linear relation between signed circle of confusion and disparity difference, AnyBokeh estimates a source-specific \textbf{optical fingerprint} and transfers the source optical characteristics to the desired focus and aperture setting. A generative editor conditioned on both source and target circle-of-confusion maps then performs relative blur synthesis, enabling spatially adaptive deblurring, preservation, and defocus rendering.
    To support physically supervised learning, we further construct a high-fidelity synthetic dataset with accurate depth, focus distance, and full EXIF metadata. Experiments on real-world benchmarks show that AnyBokeh achieves faithful and controllable editing across any-to-any bokeh editing, all-in-focus-to-bokeh rendering, and defocus deblurring, while avoiding all-in-focus reconstruction and test-time bokeh-level calibration commonly required by existing approaches.
    The code and dataset will be available at \url{https://github.com/itsmag11/AnyBokeh}.
\end{abstract}
\section{Introduction}

Depth-of-field (DoF) control is an essential aesthetic tool in photography, guiding visual attention and conveying scene structure through selective focus and optical defocus, commonly perceived as bokeh. The strength and spatial distribution of bokeh are governed by the image formation process: scene points away from the focal plane are projected as circles of confusion (CoC) whose size depends on aperture, focus distance, focal length, and scene depth. In practice, a large physical aperture, corresponding to a small f-number such as $f/2.0$, produces a shallow DoF and strong blur in out-of-focus regions, while a small physical aperture like $f/22.0$ increases the DoF and makes a wider depth range appear sharp. Although these optical effects can be controlled during capture, post-capture DoF manipulation from a single in-the-wild photograph remains highly challenging.

The difficulty mainly comes from two unknowns. First, the input may already contain spatially varying defocus blur: some regions are sharp, while others are mildly or severely out of focus. A practical editor must understand this source blur state to decide which regions should be sharpened, preserved, or further defocused. Second, the camera parameters that determine the absolute blur scale, such as focus distance, aperture, focal length, sensor size, and image resolution, are often incomplete or unavailable for in-the-wild images. Together, these two unknowns make any-to-any bokeh editing fundamentally different from standard bokeh rendering: the editor must infer the current blur state of the source image while also determining the target blur scale under the desired optical setting.

Existing methods address these two ambiguities only partially. Many bokeh rendering methods assume an all-in-focus (AIF) input~\cite{ignatov2020rendering,sheng2024drbokeh,Peng2022MPIB,Peng2022BokehMe,zhu2025bokehdiff,seizinger2025bokehlicious}, where the source blur state is trivial and only the target bokeh needs to be synthesized. This assumption simplifies the problem, but restricts these methods to already sharp inputs. 
Recent refocusing pipelines~\cite{Genfocus2025} relax the input constraint by first reconstructing an AIF image from an arbitrary input and then rendering new bokeh. We refer to this forced AIF reconstruction as an \emph{AIF bottleneck}: the input is projected into a sharp intermediate representation, which removes the observable source blur pattern and makes source-specific optical cues unavailable for subsequent target bokeh synthesis. Moreover, severe defocus deblurring is inherently ambiguous and can introduce unnecessary hallucinated details or structural errors that propagate into the final edit. 
In addition, because complete optical metadata is usually unavailable, existing pipelines often require \emph{bokeh-level calibration}~\cite{Peng2022BokehMe,zhu2025bokehdiff,Genfocus2025}, i.e., adjusting a renderer-specific blur scale to match the desired bokeh strength, through per-image ground-truth-guided search.

To address these limitations, we propose \textbf{AnyBokeh}, a physics-guided framework for any-to-any bokeh editing. Instead of treating source blur merely as a degradation to be removed, AnyBokeh explicitly estimates the source blur state as a signed source CoC map, together with a disparity map. By modeling the linear relation between signed CoC diameter and disparity difference, we estimate a source-specific CoC--disparity slope, which absorbs the global optical and scaling factors that map disparity differences to image-space defocus. We call this slope an \emph{optical fingerprint}. This quantity is physically equivalent to the bokeh-level parameter $K$ used in prior rendering methods~\cite{Peng2022BokehMe,zhu2025bokehdiff,Genfocus2025}, as both control how disparity differences are converted into CoC magnitude. In our formulation, we denote it by $\kappa_{src}$ to emphasize that this CoC scale is estimated directly from the observed source blur state and disparity, rather than specified externally. We then transfer this optical fingerprint to the desired focus and aperture setting to obtain the target CoC map, avoiding AIF reconstruction and test-time bokeh-level calibration.

AnyBokeh synthesizes the target image with a generative editor conditioned on the source image, source CoC, and target CoC. We refer to this design as \emph{dual-CoC conditioning}: both the current and desired blur states are provided to the model, turning any-to-any bokeh editing into relative blur synthesis. This allows the editor to explicitly model the transition from source blur to target blur and perform spatially adaptive bokeh synthesis. To train the CoC and disparity estimator under accurate physical supervision, we further construct \textbf{UnrealBokeh}, a high-fidelity synthetic dataset rendered with ground-truth depth, focus distance, aperture, and focal length.
The contributions of AnyBokeh are summarized as follows:
\begin{itemize}
    \item We propose a CoC-mediated optical fingerprint transfer framework for any-to-any bokeh editing. By estimating source CoC and disparity, our method extracts a source-specific CoC--disparity slope $\kappa_{src}$ and analytically transfers it to the target focus and aperture setting, avoiding AIF reconstruction and test-time bokeh-level calibration.
    \item We introduce dual-CoC conditioning for relative blur synthesis. By conditioning the editor on both source and target CoC, our model explicitly observes the input blur state and the desired output blur state, enabling spatially selective deblurring, preservation, and defocus synthesis.
    \item We build a high-fidelity synthetic dataset for physically supervised CoC learning. Using Unreal Engine, we render diverse scenes with ground-truth depth, focus distance, aperture, and focal length across multiple optical configurations, enabling robust learning of source CoC and disparity estimation.
\end{itemize}
\section{Related Work}
\label{sec:related}

\noindent\textbf{Bokeh Rendering.}
Classical depth-of-field rendering traditionally relies on physically based simulations, which produce realistic bokeh but are computationally expensive and demand complete 3D geometry~\cite{potmesil1981lens,haeberli1990accumulation}. To improve efficiency, image-space post-filtering methods have been widely adopted~\cite{kraus2007depth,lee2010realtime,scattering,barron2015fast,peng2021interactive}. However, they inherently struggle with depth discontinuities, causing severe color bleeding and partial occlusion artifacts~\cite{busam2019sterefo,sheng2024drbokeh}. While subsequent approaches attempted to mitigate these issues via physical lens models, optical aberrations, or multiview and light-field synthesis~\cite{hach2015cinematic,wu2012rendering,lee2009depth,yu2010realtime}, handling complex occlusions remains a persistent challenge. To fundamentally resolve these geometric artifacts, recent research has shifted towards layered and neural rendering paradigms.
Early deep models and benchmarks study natural camera bokeh rendering from all-in-focus images~\cite{ignatov2020rendering}, while later methods combine depth estimation, layered scene representations, differentiable rendering, or neural priors to improve photorealism and boundary handling. MPIB~\cite{Peng2022MPIB} leverages multiplane images and background inpainting to address high-resolution occlusions. BokehMe~\cite{Peng2022BokehMe} and BokehMe++~\cite{bokehme++} combine classical filtering with neural refinement to reduce boundary artifacts. Dr.~Bokeh~\cite{sheng2024drbokeh} introduces a differentiable occlusion-aware renderer, and Bokehlicious~\cite{seizinger2025bokehlicious} removes the need for explicit depth estimation by learning aperture-aware attention. More recently, the generative priors of diffusion models have been leveraged to overcome the limitations of traditional neural networks. Methods like Bokeh Diffusion~\citep{fortes2025bokehdiffusion} and Generative Photography~\cite{Yuan_2024_GenPhoto} explicitly inject bokeh conditions into text-to-image synthesis, while BokehDiff~\citep{zhu2025bokehdiff} adapts a one-step diffusion scheme for high-fidelity neural lens blur.
Despite significant advancements in synthesis quality and computational efficiency, these rendering-based architectures strictly demand AIF images as inputs. 

To break the AIF input limitation and achieve arbitrary refocusing, recent frameworks like DiffCamera~\citep{wang2025diffcamera} propose a DiT-based~\cite{Peebles2023DiT} framework trained on large-scale simulated depth-of-field image pairs. It introduces a physics-inspired stacking constraint and a depth dropout mechanism to enable flexible control over arbitrary focal points and bokeh levels. However, we argue that their method may establish potential exposure bias, such that the defocus maps used in bokeh-to-clear rendering are from clear images, leading to information leak. Genfocus~\cite{Genfocus2025}, on the other hand, adopts a cascaded two-stage pipeline. It first employs a DeblurNet to force the recovery of an AIF image from an arbitrary input, and subsequently utilizes a BokehNet trained via semi-supervised learning on both synthetic and real-world data to render the target bokeh with controllable focus and blur intensity. However, by strictly decoupling the process into isolated deblurring and re-blurring stages, Genfocus prematurely discards the original optical context and introduces unnecessary hallucination.

\noindent\textbf{Defocus Deblurring}
The core objective of defocus deblurring is to recover a sharp AIF image from inputs degraded by spatially varying depth-of-field blur. Unlike uniform motion blur, defocus blur is tightly coupled with scene depth and camera aperture, and its magnitude can change abruptly across depth discontinuities and occlusion boundaries. Early methods typically formulated the problem as spatially varying deconvolution with estimated defocus maps or blur kernels~\cite{bando2007towards,zhu2013estimating}, but such pipelines are sensitive to inaccurate blur estimation and often introduce artifacts around complex depth transitions. 
Recent learning-based methods improve robustness by introducing stronger image priors and defocus-specific cues. Some methods exploit dual-pixel~\citep{abuolaim2020defocus, pan2021dual} or quad-pixel~\cite{chen2025quadpixel} sensors to obtain aperture-dependent sub-view information, while others model defocus through adaptive filters~\cite{Lee2021IFAN}, disparity-aware kernels~\cite{yang2023k3dn}, or implicit inverse kernels~\cite{quan2023single,tang2024prior}. These works show that effective defocus deblurring requires not only generic image restoration capacity, but also an understanding of the relationship between blur magnitude and scene geometry.
Following this observation, we compare against two representative baselines. DRBNet~\citep{ruan2022drbnet} learns the deblurring mapping by jointly leveraging light-field-generated bokeh and real defocus images. However, when applied to highly challenging, in-the-wild photographs with severe blur, it frequently fails to recover the underlying geometric structures, leading to significant structural distortions in the output. Conversely, Restormer~\citep{Zamir2021Restormer} is a highly efficient Transformer architecture tailored for high-resolution image restoration. Despite its strong general performance in denoising and deblurring, its inherently regressive objective struggles with the severely ill-posed nature of extreme defocus, often yielding overly smoothed, averaged textures rather than authentic high-frequency details. Furthermore, it is prone to introducing unnatural artifacts in background regions with complex depth variations.

\section{Methodology}
\label{sec:method}

\subsection{Preliminary}
\label{sec:background}

\textbf{Circle-of-Confusion.}
Under the thin-lens model with a circular aperture, a scene point away from the focal plane is imaged as a defocus disk, commonly referred to as the circle of confusion (CoC). Let $N$ denote the f-number, $f$ the focal length, $S_1$ the focus distance, and $S_2$ the scene-point distance. All distances are measured in millimeters, while $N$ is dimensionless. The signed physical CoC diameter is given by
\begin{equation}
    CoC^{mm} = \frac{f^2}{N(S_1 - f)} \frac{(S_2 - S_1)}{S_2}\,,
    \label{equ:1}
\end{equation}
The sign indicates whether the scene point lies in front of or behind the focal plane, while the magnitude gives the blur diameter. Throughout the paper, we use $CoC$ to denote signed CoC diameter unless otherwise specified.

To express the CoC in image pixels ($px$), the physical blur diameter is converted according to the sensor size and image resolution:
\begin{equation}
    CoC^{px} = CoC^{mm} \cdot \frac{W_{res}}{W_{sensor}}\,,
    \label{equ:2}
\end{equation} 
where $W_{res}$ is the image width in $px$ and $W_{sensor}$ is the physical sensor width in $mm$. Note that any image resizing changes the CoC magnitude in pixel units.
In our Unreal Engine~\cite{unrealengine} synthetic dataset (Sec.~\ref{sec:unrealbokeh}), the per-pixel scene depth, focus distance, focal length, f-number, sensor size, and image resolution are all available. We therefore compute dense ground-truth signed CoC maps paired with each rendered RGB image, providing direct supervision for learning CoC estimation.

\textbf{Optical Fingerprint.}
Eq.~\ref{equ:1} shows that the CoC is determined by both scene geometry and camera optics. In typical non-macro photography, the focus distance is much larger than the focal length, i.e., $S1 \gg f$. Thus, $S_1 - f \approx S_1$, and the signed CoC in physical unit can be approximated as
\begin{equation}
\begin{split}
    CoC^{mm} & \approx \frac{f^2}{N \cdot S_1} \frac{(S_2 - S_1)}{S_2} = \frac{f^2}{N}\cdot(\frac{1}{S_1}-\frac{1}{S_2})\,,
    \label{equ:3}
\end{split}
\end{equation}
Let $D = 1/S_2$ denote the disparity of a scene point and $D_{focus} = 1/S_1$ denote the focus disparity. Substituting the above approximation into Eq.~\ref{equ:2}, the pixel-space CoC becomes
\begin{equation}
    CoC^{px} = \underbrace{
    \left(
    \frac{f^2 \cdot W_{res}}{N \cdot W_{sensor}}
    \right)
    }_{\kappa}
    \cdot (D_{focus} - D) \,.
\label{equ:slope}
\end{equation}
This formulation reveals a linear relationship between signed CoC and disparity difference, where the slope $\kappa$ is shared by all pixels of the same source image. Therefore, even when EXIF metadata is unavailable or incomplete, the slope of this relationship can be estimated from the predicted source CoC and disparity. It absorbs the global factors that determine the CoC scale, including aperture, focal length, sensor size, image resolution, and the scale convention of the estimated disparity. We refer to this source-specific slope as the \textit{optical fingerprint}, since it characterizes how the camera configuration and depth representation convert disparity differences into image-space defocus.

\begin{figure}
    \centering
    \includegraphics[width=\linewidth]{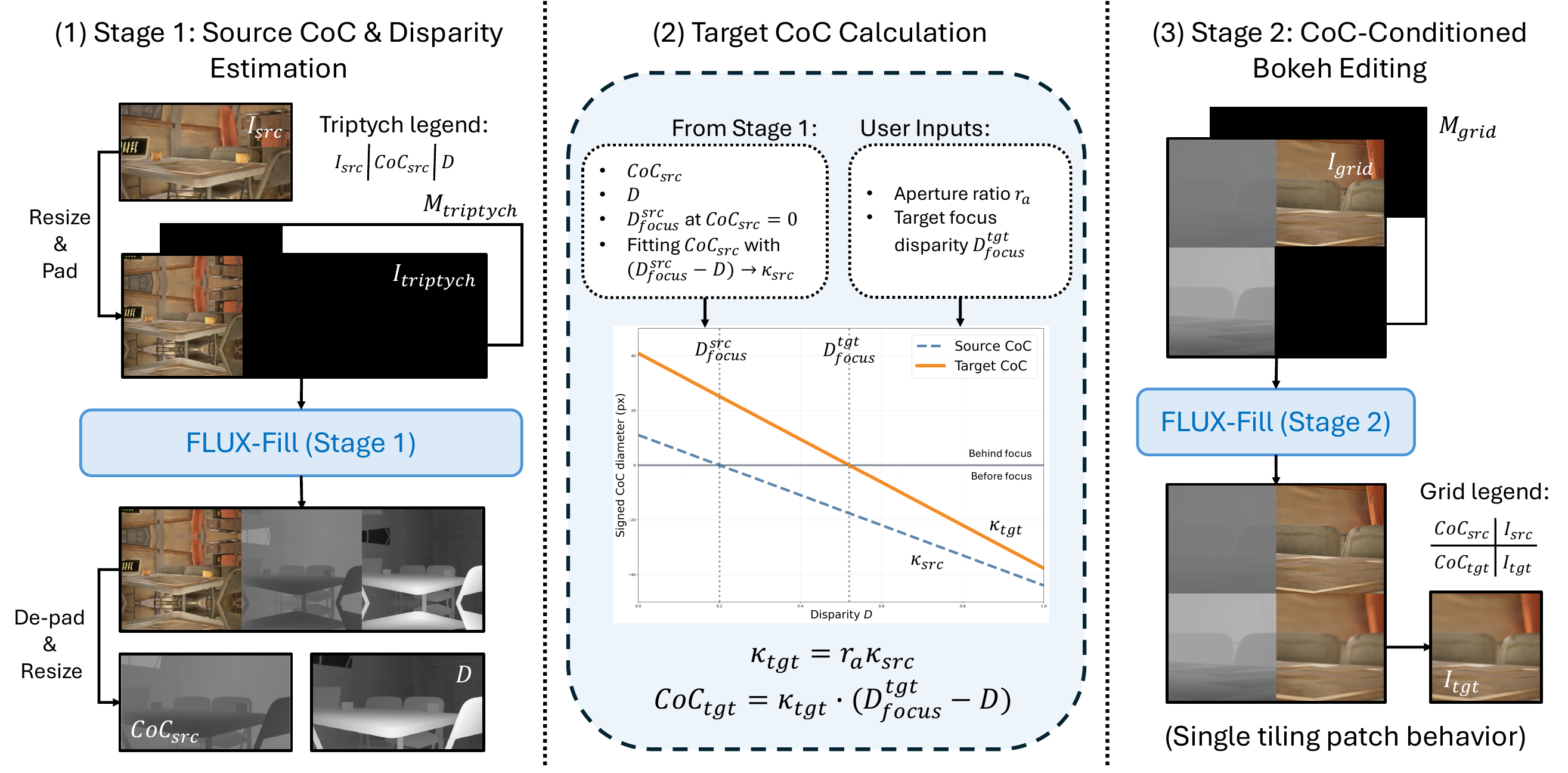}
    \caption{Overview of the AnyBokeh framework. (1) Stage~1 predicts the signed source CoC map $CoC_{src}$ and disparity map $D$ from the source image to recover the source blur state and geometry. (2) The target CoC is analytically computed by transferring the source optical fingerprint $\kappa_{src}$ to the desired focus and aperture setting. (3) Stage~2 performs CoC-conditioned bokeh editing using the source image, $CoC_{src}$, and $CoC_{tgt}$ to synthesize the final target image.}
    \label{fig:framework}
\end{figure}

\subsection{Stage 1: CoC and Disparity Estimation}
\label{sec:stage1}

Stage~1 of our method estimates the source blur state and geometry from a single input image. Given the source image $I_{src}$, it jointly predicts two dense maps: the signed source CoC map $CoC_{src}$ and the disparity map $D$. The source CoC describes the defocus already present in the input image, while disparity provides the geometric coordinate needed to transfer the optical fingerprint to a new focus setting. Predicting these two quantities together encourages the model to reason about defocus and geometry in a coupled manner: $CoC_{src}$ provides direct evidence of the observed blur state, while $D$ explains the scene geometry underlying that blur. 

Estimating $CoC_{src}$ and $D$ from defocused inputs is inherently ambiguous, especially around depth discontinuities and severely blurred structures where high-frequency evidence has been weakened. We therefore rely on a generative prediction model with a strong image prior, which can infer plausible scene boundaries and coherent geometry from degraded visual clues. This prior is particularly useful for producing sharp and spatially consistent dense maps, rather than overly smooth estimates that would lead to inaccurate CoC--disparity slopes. The CoC map is normalized as detailed in Sec.~\ref{sec:exp}.

We predict $CoC_{src}$ and $D$ on the full resized image rather than using tiled inference. This is important because disparity is defined up to a global scale convention, and independently predicting different tiles may lead to inconsistent disparity scales or discontinuities across tile boundaries. Such inconsistencies would directly affect the estimated CoC--disparity slope and therefore the optical fingerprint. Instead, Stage~1 predicts globally consistent low-resolution maps from the complete image context, which are then de-padded and upsampled to the original resolution.
This formulation is agnostic to the unknown camera metadata or prior image transformations of an in-the-wild photo. Stage~1 does not require knowing how the image was captured or resized before being observed. It simply estimates the signed CoC diameter visible on the current input canvas. As a result, the predicted $CoC_{src}$ and $D$ provide a consistent basis for estimating the source optical fingerprint. The framework of Stage~1 is shown in Fig.~\ref{fig:framework}~(1).

\noindent\textbf{Target CoC Calculation.}
Given the source CoC and disparity from Stage~1, our goal is to construct a target CoC map that specifies the desired blur state for Stage~2. We primarily consider the relative editing setting, where the source image provides observable defocus cues and the user specifies the desired change in focus and aperture. This setting is suitable for in-the-wild bokeh editing, since complete EXIF metadata is often unavailable, but the source blur pattern still carries information about the underlying optical scale. 

As derived in Eq.~\ref{equ:slope}, the signed CoC is approximately linear with respect to the disparity difference $(D_{focus}^{src} - D)$. $D_{focus}^{src}$ can be directly obtained as the disparity at $CoC_{src}=0$. For a given source image, the slope $\kappa_{src}$ captures the source-specific CoC scale and serves as the optical fingerprint that both source and target images share. Once $\kappa_{src}$ is estimated from the predicted $CoC_{src}$ and $D$, we transfer it to the target setting by scaling it according to the desired aperture change. Let $r_a$ denote the aperture ratio from the source to the target setting, e.g., $r_a = N_{src} / N_{tgt}$ when the focal length and sensor scale are fixed. The target CoC--disparity slope and target CoC are computed as
\begin{equation}
    \kappa_{tgt} = r_a \kappa_{src}\,, \qquad
    CoC_{tgt} = \kappa_{tgt}\left(D^{tgt}_{focus} - D\right)\,,
\label{equ:target_coc}
\end{equation}
where $D^{tgt}_{focus}$ is the user-specified target focus disparity. An illustration can be found in Fig.~\ref{fig:framework}~(2). This formulation transfers the optical scale observed in the source image to the desired focus and aperture setting without requiring complete camera metadata. Unlike previous methods~\cite{Peng2022BokehMe, zhu2025bokehdiff,Genfocus2025}, the target CoC is obtained directly from the source optical fingerprint and user control, avoiding test-time bokeh-level calibration. Moreover, since the target CoC is derived by scaling the source optical fingerprint, the predicted blur strength remains consistent with the defocus scale observed in the input image, rather than being determined by an independently searched parameter.

The proposed AnyBokeh also supports absolute CoC control by directly specifying $\kappa_{abs}$, which recovers the standard bokeh rendering form:
\begin{equation}
    CoC_{tgt} = \kappa_{abs} \left(D^{tgt}_{focus} - D\right).
\end{equation}
This setting is consistent with conventional bokeh rendering pipelines, which synthesize target defocus from an AIF input using an externally specified blur strength. 
Since such methods do not preserve or estimate the source optical fingerprint, absolute control is their primary way to determine the target CoC scale, often requiring manual tuning or test-time search in practice. 
In contrast, AnyBokeh uses absolute control only as an optional interface, e.g., when users want to impose a specific absolute target blur scale or when the source defocus cues are too weak to yield a reliable fingerprint. However, an ideal all-in-focus image with exactly zero CoC is rare in practice, and the predicted source CoC typically provides a usable, though sometimes weak, slope for relative transfer.

\subsection{Stage 2: CoC-Conditioned Bokeh Editing}

Given the source image $I_{src}$, $CoC_{src}$, and $CoC_{tgt}$, Stage~2 learns the mapping $(I_{src}, CoC_{src}, CoC_{tgt}) \rightarrow I_{tgt}$. Unlike conventional AIF-to-bokeh rendering~\cite{ignatov2020rendering,sheng2024drbokeh,Peng2022MPIB,Peng2022BokehMe,zhu2025bokehdiff,seizinger2025bokehlicious}, our setting handles inputs that may already contain defocus blur. Therefore, the model needs to determine not only where to add blur, but also where to preserve existing content or recover sharper structures.
To this end, we use \emph{dual-CoC conditioning}: both $CoC_{src}$ and $CoC_{tgt}$ are provided to the editor as explicit blur-state guidance. The source CoC describes the blur already present in the input image, while the target CoC specifies the desired blur distribution after editing. Together, they define a relative blur transition, allowing the editor to reason about whether each region should be deblurred, preserved, or further defocused. Both CoC maps are normalized as detailed in Sec.~\ref{sec:exp}.
This design is important for any-to-any editing because the target CoC alone does not reveal the source blur state. If $CoC_{src}$ is omitted, the model must infer the existing defocus solely from image appearance, which is ambiguous and can lead to inaccurate bokeh synthesis (see Sec.~\ref{sec:ablation}).
Combined with the target CoC calculation in Sec.~\ref{sec:stage1}, Stage~2 completes the any-to-any editing pipeline by turning the transferred optical fingerprint into the final refocused image. The framework is shown in Fig.~\ref{fig:framework}~(3).

\subsection{UnrealBokeh Synthetic Dataset}
\label{sec:unrealbokeh}

\begin{figure}[t]
    \centering
    \includegraphics[width=\linewidth]{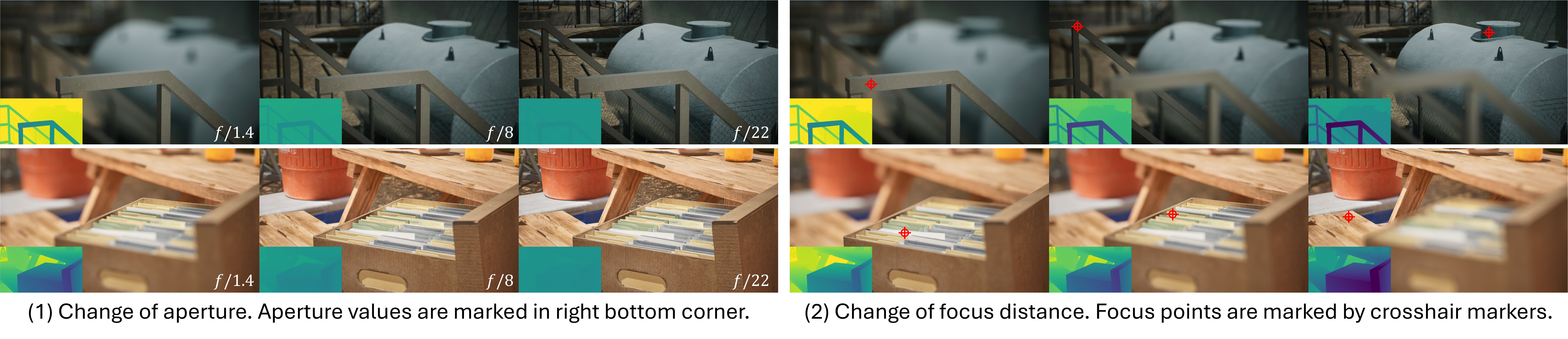}
    \caption{Samples from UnrealBokeh. For the same (environment, camera, focal length) scene, images with various apertures and focus distances are captured. Sharp ground truth CoC maps derived from accurate depth maps are provided, as demonstrated in the bottom left corner of each image.}
    \label{fig:ub}
\end{figure}

Training Stage~1 requires dense supervision for both source CoC and disparity estimation. Accurate CoC supervision depends not only on scene depth, but also on optical parameters such as focus distance, focal length, aperture, sensor size, and image resolution. Such annotations are difficult to obtain from real photographs, especially for in-the-wild images where complete metadata, depth, and focus annotations are typically unavailable.
Existing bokeh datasets are not sufficient for this purpose. Real datasets such as EBB!~\cite{ignatov2020rendering} and RealBokeh~\cite{seizinger2025bokehlicious} provide paired images captured under different camera settings, but do not include the dense geometry and complete optical metadata needed to compute ground-truth CoC maps. Synthetic datasets generated by classical ray-tracing~\cite{Peng2022BokehMe,zhu2025bokehdiff} provide scalable training data, but often rely on simplified layer-based geometry that cannot fully represent complex scene depth.

To enable physically supervised CoC learning, we construct \textbf{UnrealBokeh}, a high-fidelity synthetic dataset rendered in Unreal Engine~\cite{unrealengine}. For each image, we record RGB, dense depth, focus distance, aperture, focal length, sensor size, and image resolution, allowing us to compute dense signed CoC maps from the thin-lens model. Sample images with corresponding CoC maps from UnrealBokeh are shown in Fig.~\ref{fig:ub}. We render diverse indoor and outdoor scenes under controlled optical variations and form source--target editing pairs by varying focus distance and aperture while keeping scene geometry and camera viewpoint fixed. The resulting dataset contains 11{,}477 images and 382{,}888 editing pairs. Comparisons with existing datasets are listed in Tab.~\ref{tab:dataset_comparison}. We split the data by environment--camera group to avoid scene and viewpoint leakage across training and validation. UnrealBokeh provides physically grounded supervision for Stage~1 and aligned source--target pairs for training Stage~2. Additional dataset construction details are provided below.

\textbf{Scene Collection.}
UnrealBokeh is rendered using Unreal Engine with diverse indoor and outdoor 3D scenes. We download 3D environments from the Fab marketplace of Unreal Engine~\citep{unrealengine} and obtain authentication from the authors for photo rendering in their 3D assets for image-level research use. These environments provide realistic geometry, materials, and lighting, enabling more faithful optical supervision than simplified layer-based synthetic data.

\textbf{Rendering Setup.}
We instantiate 71 unique camera setups across the collected environments. For each camera setup, we render images under multiple focal lengths selected from $\{11, 24, 35, 55, 85, 135\}$ mm with a small random perturbation of $\pm 10$ mm. For each focal length, we vary the f-number from $f/1.4$ to $f/22.0$ and render images with different focus distances. For every rendered image, we store the RGB image, dense depth map, focus distance, aperture, focal length, sensor size, and image resolution.

\textbf{Ground-truth CoC Computation.}
Using the stored depth and camera parameters, we compute dense signed CoC maps following the thin-lens formulation in Sec.~\ref{sec:background}. The sign of CoC indicates whether a scene point lies in front of or behind the focal plane, and the magnitude denotes the image-space blur diameter. These dense signed CoC maps provide direct supervision for Stage~1 source CoC prediction.

\textbf{Pair Construction.}
To construct source--target editing pairs, we group rendered images by environment, camera, and focal length. Within each group, images share the same scene geometry and viewpoint but differ in focus distance and aperture. We pair images from the same group to form aligned bokeh editing pair examples. Pairs with overly similar source and target appearances are removed to improve training diversity. This process results in 11{,}477 rendered images and 382{,}888 unique source--target editing pairs.

\textbf{Data Split.}
We split UnrealBokeh into training and validation sets by environment--camera group, resulting in 340{,}796 editing pairs in the train set and 42{,}092 pairs in the validation. This ensures that images from the same scene and viewpoint do not appear across different splits, preventing leakage between training and validation. 
We use UnrealBokeh to train Stage~1 with physically accurate CoC and disparity supervision, and to train Stage~2 with aligned source--target editing pairs.

\begin{table}[htbp!]
    \centering
    \caption{Dataset comparison. UnrealBokeh provides complete optical supervision and dense geometry, enabling ground-truth CoC computation for Stage~1 training. 
    *Actual downloadable resolution of RealBokeh is different from $6000{\times}4000$ stated in the paper~\cite{seizinger2025bokehlicious}.
    }
    \label{tab:dataset_comparison}
    \setlength{\tabcolsep}{5pt}
    \renewcommand{\arraystretch}{0.95}
    \begin{tabular}{lccc}
    \toprule
    Dataset & \textbf{UnrealBokeh} & RealBokeh~\cite{seizinger2025bokehlicious} & EBB!~\cite{ignatov2020rendering} \\
    \midrule
    \# Images & 11,477 & 27,451 & 9,774 \\
    \# Editing Pairs & 382,888 & 224,750 & 9,774 \\
    Resolution & $2016{\times}1120$ & $2000{\times}1500$* & $\approx1550{\times}1024$ \\
    Aperture Range & $f/22$--$f/1.4$ & $f/20$--$f/2$ & $f/16$,$f/1.8$ \\
    \midrule
    Multiple Focus Distances & \cmark & \xmark & \xmark \\
    Complete EXIF Metadata & \cmark & \xmark & \xmark \\
    Dense GT Depth & \cmark & \xmark & \xmark \\
    Computable Dense GT CoC & \cmark & \xmark & \xmark \\
    \bottomrule
    \end{tabular}
\end{table}
\section{Experiments}
\label{sec:exp}

\begin{figure}[t]
    \centering
    \includegraphics[width=0.9\linewidth]{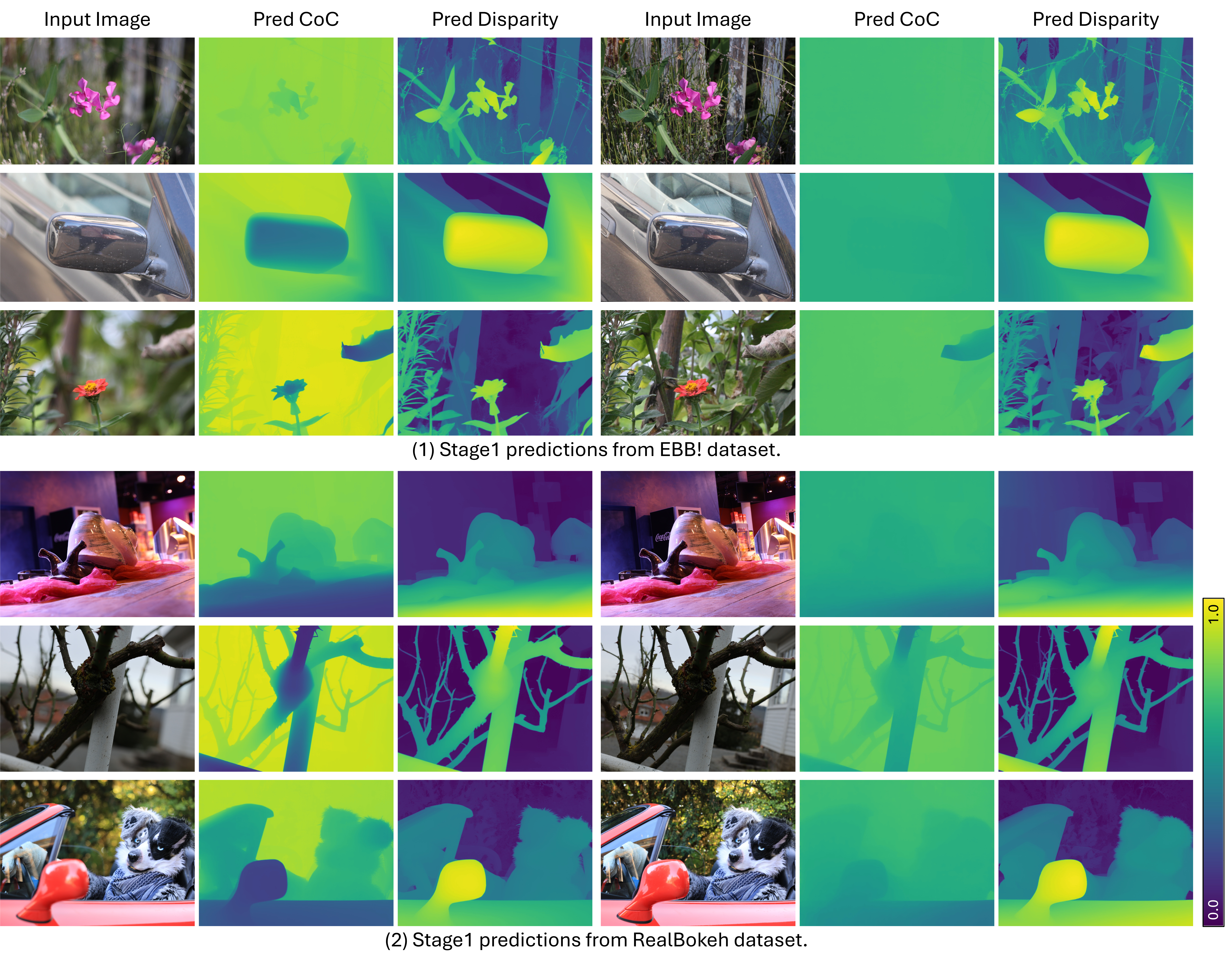}
    \caption{Qualitative Stage~1 predictions on EBB! and RealBokeh. Each row shows the same scene under two aperture settings, with stronger defocus on the left and weaker defocus on the right. Stage~1 predicts larger CoC magnitudes for stronger bokeh while producing consistent disparity maps across aperture changes, indicating that it can separate source blur state from scene geometry.}
    \label{fig:stage1}
\end{figure}

\begin{figure}
    \centering
    \includegraphics[width=\linewidth]{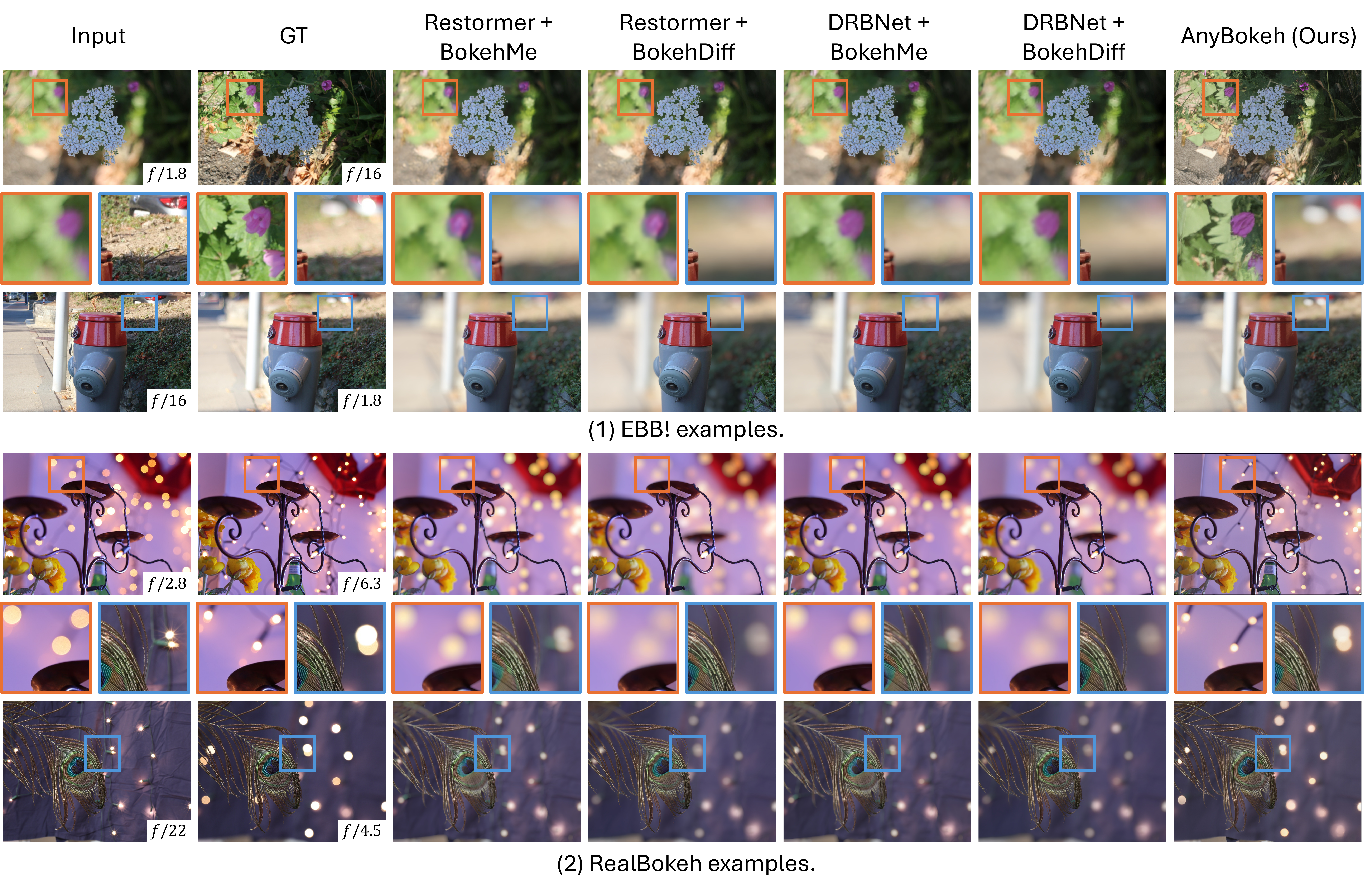}
    \caption{Qualitative comparison on any-to-any bokeh editing. Aperture values are shown for the input and ground-truth target images. The top row of each dataset shows bokeh removal, and the bottom row shows bokeh enhancement. Compared with deblur-then-render baselines, AnyBokeh performs spatially appropriate deblurring or defocus synthesis more effectively, adjusting bokeh balls faithfully by enlarging or shrinking them according to the aperture changes.}
    \label{fig:any2any}
\end{figure}

\textbf{Implementation Details.}
For both Stage~1 and Stage~2, we use FLUX.1-Fill-dev\footnote{black-forest-labs/FLUX.1-Fill-dev}~\cite{flux2024} as the generative backbone and fine-tune it with LoRA~\citep{hu2022lora}. We set the LoRA rank to 32 and train each stage for approximately 30K iterations with a per-GPU batch size of 1. Learning rate is fixed at $1e-4$ with AdamW optimizer~\cite{loshchilov2018adamw}. Training of both stages is done on 4 NVIDIA A100 80 GB GPUs. Offset noise is applied during fine-tuning of both stages.

For Stage~1, we first resize the longer side of the source image to the model input resolution and apply reflection padding to obtain a square input. The raw CoC map is normalized to $[0,1]$ by logarithmic normalization. We formulate the prediction as an inpainting task on a horizontal triptych canvas, where the left panel contains the source image $I_{src}$, the middle panel corresponds to the source CoC map $CoC_{src}$, and the right panel corresponds to the disparity map $D$. During inference, the source image panel is kept visible, while the CoC and disparity panels are masked as the inpainting regions, as illustrated in Fig.~\ref{fig:framework}. 
The model then fills the two masked panels to produce image-like predictions of $CoC_{src}$ and $D$.
After prediction, we extract the CoC and disparity panels, remove the padding, and upsample both maps back to the original image resolution. Since CoC is defined as an image-space blur diameter in pixels, its magnitude must be rescaled consistently with the image resizing factor. Specifically, if the source image is resized by a factor $s$ before Stage~1 prediction, the recovered CoC map is scaled by $1/s$ when mapped back to the original resolution. The CoC map is also denormalized to its original raw value range. The disparity map is spatially resized, but its values are left unchanged, since it serves as a relative geometric coordinate. Target CoC is calculated in raw value range.

For Stage~2, we use signed $\tanh$ normalization for $CoC_{src}$ and $CoC_{tgt}$ to represent arbitrary raw CoC values in a bounded image-like form. The input is arranged as a $2\times2$ grid as shown in Fig.~\ref{fig:framework}: the top-left cell contains $CoC_{src}$, the top-right cell contains the source image $I_{src}$, the bottom-left cell contains $CoC_{tgt}$, and the bottom-right cell is the target image slot $I_{tgt}$. During training, we sample aligned image patches and construct one such grid for each patch. The target image slot is masked, and the model learns to synthesize $I_{tgt}$ conditioned on the other three cells. During inference, we apply the same grid formulation to overlapping tiles of the full-resolution image. The predicted target tiles are merged using Hann-windowed overlap-add stitching with a window overlap of $0.2$, preserving the original resolution and aspect ratio.

\textbf{CoC Map Normalization.}
We use different CoC normalization strategies for Stage~1 and Stage~2 since CoC plays different roles in the two stages. In Stage~1, CoC is a regression target and must be predicted accurately. In Stage~2, CoC is used as a conditioning signal for image generation and mainly needs to provide stable blur-state guidance.

For Stage~1, directly regressing raw pixel-space CoC values is suboptimal because CoC maps have a large dynamic range. Large out-of-focus regions can dominate the regression target, while small CoC errors near the focal plane are often more important for refocusing control, since they can shift the perceived focus boundary. We therefore use a signed logarithmic normalization:
\begin{equation}
    CoC_{signed}^{log}
    =
    \operatorname{sign}(CoC)
    \frac{
        \log(1 + |CoC|)
    }{
        \log(1 + CoC_{max})
    },
    \qquad
    CoC_{norm}^{log}
    =
    \operatorname{clip}
    \left(
        \frac{CoC_{signed}^{log} + 1}{2},
        0, 1
    \right).
\end{equation}

This transformation maps the raw CoC range $[-CoC_{max}, CoC_{max}]$ to $[0,1]$, with zero CoC mapped to $0.5$. It allocates more representational value range to near-focus regions, where accurate CoC prediction is most critical, while still preserving distinctions among large CoC values. This is important for Stage~1 because the model is explicitly supervised to predict CoC values, and over-saturating large CoC regions would weaken the learning signal for heavily defocused areas. 

For Stage~2, CoC maps are not regression targets but conditioning inputs. The editor only needs them to describe the source and target blur states in a stable and bounded form. We therefore use a signed $\tanh$ normalization:
\begin{equation}
    CoC_{signed}^{tanh}
    =
    \tanh
    \left(
        \frac{CoC}{S}
    \right),
    \qquad
    CoC_{norm}^{tanh}
    =
    \frac{CoC_{signed}^{tanh} + 1}{2}.
\end{equation}

where $S$ controls the saturation scale and is set as $S=0.1\times\max(H,W)$. This maps raw CoC values from $(-\infty,\infty)$ to a bounded range $[0,1]$ while preserving the CoC sign. Unlike Stage~1, saturation is acceptable here because CoC is used as guidance rather than a supervised output. The bounded representation improves numerical stability and allows the editor to handle arbitrary target CoC magnitudes, while still preserving sufficient information for the model to distinguish relative blur states.

We avoid using a $\tanh$ normalization for Stage~1 regression because it quickly saturates for large raw CoC values. Once saturated, different large CoC magnitudes become nearly indistinguishable in the normalized space, giving the model little incentive to predict large blur values accurately. This can lead to systematically underestimated CoC in strongly defocused regions.

\textbf{Evaluation Setup.}
We evaluate AnyBokeh under three settings: any-to-any bokeh editing, AIF-to-bokeh rendering, and defocus deblurring (bokeh-to-AIF). In the any-to-any setting, the source and target images are captured from the same scene and viewpoint but under different focus distances and aperture values; the source image may therefore already contain spatially varying defocus. In the AIF-to-bokeh setting, the source image is all-in-focus, and the target image contains stronger defocus, corresponding to the conventional bokeh rendering task. In the defocus deblurring setting, the source image is defocused, and the target image is all-in-focus. 
Following prior bokeh rendering methods~\citep{Peng2022BokehMe,zhu2025bokehdiff,seizinger2025bokehlicious,Genfocus2025}, we use LPIPS~\citep{lpips} and DISTS~\citep{dists} to measure reference-based perceptual similarity, and FID~\cite{fid} for distribution-level realism.

\textbf{Datasets.}
Although AnyBokeh is trained only on the synthetic UnrealBokeh dataset, we evaluate it on real-world bokeh datasets to assess zero-shot generalization. We use EBB!~\cite{ignatov2020rendering} and RealBokeh~\cite{seizinger2025bokehlicious} following \cite{Peng2022BokehMe,zhu2025bokehdiff,seizinger2025bokehlicious}. For EBB!, we use the Val294 split and construct evaluation pairs for all three tasks. For RealBokeh, we randomly sampled 300 pairs from the entire test set for any-to-any evaluation. For both AIF-to-bokeh and bokeh-to-AIF tasks, we use valid pairs from the first 50 test IDs. For all settings, source--target pairs are selected from the same scene and viewpoint to ensure that the evaluation focuses on blur editing rather than geometric misalignment.

\textbf{Baselines.} 
We compare AnyBokeh with representative open-source methods for bokeh rendering and defocus deblurring. For bokeh rendering from AIF inputs, we evaluate BokehMe~\citep{Peng2022BokehMe} and BokehDiff~\citep{zhu2025bokehdiff}, as both methods provide controllable defocus synthesis from AIF images and are trained with synthetic bokeh supervision. Depth maps required by both methods are generated by Depth-Pro~\cite{depthpro}. Focus points estimated by foreground masks~\cite{zheng2024birefnet} or edge difference are shared by all methods, including ours.
For defocus deblurring, we implement Restormer~\citep{Zamir2021Restormer} with its single-image defocus deblurring weights for the specific task, and DRBNet~\citep{ruan2022drbnet}. 
Since existing bokeh rendering methods do not natively support arbitrary defocused inputs, we construct deblur-then-render baselines for any-to-any editing by pairing an AIF estimator with a bokeh module. The AIF estimator first predicts an all-in-focus image from the source input, and the bokeh module then renders the target defocus effect. This evaluation protocol represents the natural way to extend AIF-based bokeh rendering methods to any-to-any editing, and directly contrasts with AnyBokeh's CoC-conditioned relative editing formulation. We exclude Genfocus~\cite{Genfocus2025} from quantitative comparison because its released weights are trained on our evaluation benchmarks, and its training code is unavailable. Using the released model would make the pure zero-shot comparison unfair, while retraining it using our synthetic dataset is not feasible. We therefore discuss it only qualitatively in the related work.

\begin{table*}[t]
    \small
    \centering
    \caption{Quantitative comparison on any-to-any bokeh editing. Existing bokeh rendering methods cannot directly handle defocused inputs, so we evaluate them with deblur-then-render pipelines. Bokeh Calibration required by baselines is described in Sec.~\ref{sec:any2any}.
    Best results are marked in \colorbox{cyan!20}{\textbf{bold}}.}
    \hspace{0.5pt}
    \label{tab:any2any}
    \setlength{\tabcolsep}{3.6pt}
    \renewcommand{\arraystretch}{1.0}
    \resizebox{\textwidth}{!}{
    \begin{tabular}{llccccc ccc}
    \toprule
    \multirow{2.7}{*}{AIF Estimator} & \multirow{2.7}{*}{Bokeh Module} 
    & \multicolumn{2}{c}{Editing Protocol} 
    & \multicolumn{3}{c}{EBB!} & \multicolumn{3}{c}{RealBokeh} \\
    \cmidrule(lr){3-4}\cmidrule(lr){5-7}\cmidrule(lr){8-10}
    & & Direct Edit & No Calib. & LPIPS$\downarrow$ & DISTS$\downarrow$ & FID$\downarrow$ & LPIPS$\downarrow$ & DISTS$\downarrow$ & FID$\downarrow$ \\
    \midrule
    Restormer & BokehMe   & \xmark & \xmark & 0.4451 & 0.2634 & 38.2801 & 0.3190 & 0.1399 & 48.2732 \\
    Restormer & BokehDiff & \xmark & \xmark & 0.4449 & 0.2663 & 42.5686 & 0.3360 & 0.1505 & 61.6716 \\
    DRBNet    & BokehMe   & \xmark & \xmark & 0.4424 & 0.2588 & 36.8142 & 0.3211 & 0.1415 & 48.6855 \\
    DRBNet    & BokehDiff & \xmark & \xmark & 0.4444 & 0.2634 & 42.1294 & 0.3424 & 0.1557 & 63.3783 \\
    \midrule
    -- & \textbf{AnyBokeh (Ours)} & \cmark & \cmark & \cellcolor{cyan!20}\textbf{0.4010} & \cellcolor{cyan!20}\textbf{0.1426} & \cellcolor{cyan!20}\textbf{24.6042} & \cellcolor{cyan!20}\textbf{0.3033} & \cellcolor{cyan!20}\textbf{0.0944} & \cellcolor{cyan!20}\textbf{33.6655} \\
    \bottomrule
    \end{tabular}}
\end{table*}

\subsection{Source CoC and Disparity Map Estimation}
\label{app:stage1qual}

Fig.~\ref{fig:stage1} shows qualitative predictions of the Stage~1 estimator on EBB! and RealBokeh images. Each row contains the same scene captured with different aperture settings: the left example has a smaller f-number and stronger defocus, while the right example has a larger f-number and weaker defocus. The predicted CoC maps correctly reflect this change in bokeh strength. Specifically, inputs with stronger defocus produce larger CoC magnitudes, whereas inputs with weaker defocus yield CoC values closer to the focus value of $0.5$ in the normalized visualization. Here, $0.5$ denotes zero raw CoC, while values close to $0$ or $1$ indicate large signed defocus (as described in Sec.~\ref{sec:exp}).

The predicted disparity maps remain largely consistent across different aperture settings of the same scene, since scene geometry should be independent of the amount of optical blur. This consistency suggests that Stage~1 learns to separate geometric structure from defocus magnitude, which is essential for estimating a reliable optical fingerprint. Some ambiguity remains in regions where severe defocus removes most geometric evidence, such as background leaves in the third row, but the overall scene layout and major depth discontinuities are still recovered. These results demonstrate that Stage~1 can effectively infer both source CoC and disparity from inputs with varying bokeh levels, providing stable guidance for subsequent target CoC computation and CoC-conditioned editing.

\subsection{Any-to-Any Bokeh Editing}
\label{sec:any2any}

Any-to-any bokeh editing is the main setting of our work. As discussed above, existing AIF-to-bokeh rendering methods cannot be directly applied to defocused inputs, so we evaluate them through deblur-then-render pipelines by pairing an AIF estimator with a bokeh renderer. A remaining challenge is how to set the bokeh-level parameter used by the renderer. Since these pipelines do not estimate a source-specific optical fingerprint, their target blur scale must be provided externally.
A common strategy in prior work is to perform per-image search over the bokeh-level parameter $K$ using the ground-truth target image as guidance~\cite{Peng2022BokehMe,zhu2025bokehdiff,Genfocus2025}. However, such target-guided search is unavailable in blind post-capture editing and would make the comparison overly optimistic. Instead, we estimate a metadata-derived pseudo CoC-diameter scale following Bokeh Diffusion~\cite{fortes2025bokehdiffusion}:
\begin{equation}
    K_{pseudo} \approx \frac{f^2 S_1}{N(S_1 - f)} \cdot \frac{W_{res}}{W_{sensor}}\,,
\end{equation}
where the notation follows Sec.~\ref{sec:background}. In practice, $K_{pseudo}$ may still differ from the effective bokeh-level parameter required by each renderer due to inaccurate depth estimation, missing or imprecise metadata, image resizing, and renderer-specific blur conventions. We therefore introduce a single global calibration scale $\gamma$ and set
\begin{equation}
    K_{calib} = \gamma K_{pseudo}\,.
\end{equation}
$K_{calib}$ corresponds to the bokeh-level parameter that prior methods search for. For each method--dataset combination, we search per-image $\gamma$ once on a subset of 100 sampled image pairs and then fix the average as the optimal global scale for all test images. This protocol avoids per-image target-guided search while still allowing each baseline to use a reasonable global bokeh-level calibration.

\begin{figure}[t]
    \centering
    \includegraphics[width=0.9\linewidth]{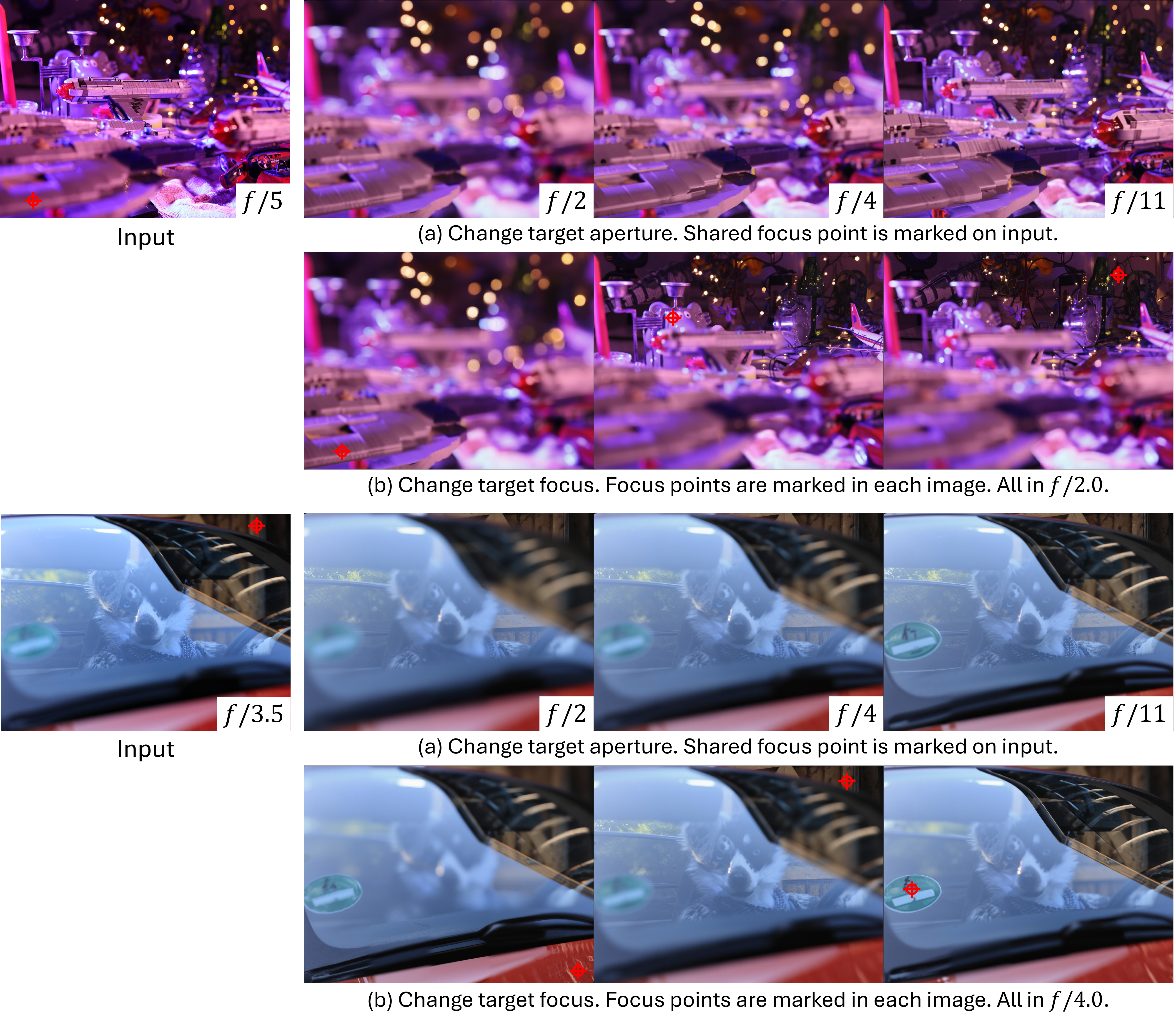}
    \caption{Qualitative results for arbitrary aperture and focus changes. Given a single input image, AnyBokeh supports user-specified aperture rendering and focus relocation without ground-truth target images. In aperture-control examples, the focus point is fixed (marked in the input), and the target f-number is varied. In focus-control examples, the aperture is fixed, and the focus point is moved to different scene locations (marked in each result image), causing the sharp region to shift accordingly.}
    \label{fig:demo}
\end{figure}

AnyBokeh does not require this calibration. Its target CoC is computed by transferring the source optical fingerprint estimated from $CoC_{src}$ and $D$ (visual examples shown in Fig.~\ref{fig:stage1}), so the target blur scale is derived from the input image and the user-specified focus/aperture change. As shown in Tab.~\ref{tab:any2any}, AnyBokeh outperforms the deblur-then-render baselines across both EBB! and RealBokeh. These results suggest that directly modeling the source and target blur states is more effective than first converting the input into an AIF image. Moreover, AnyBokeh achieves this without any test-time bokeh-level calibration, demonstrating the benefit of CoC-mediated optical fingerprint transfer for blind any-to-any editing. Visual results are showcased in Fig.~\ref{fig:any2any} and \ref{fig:demo}.

\textbf{Arbitrary Aperture and Focus Changes.}
We further demonstrate the controllability of AnyBokeh under user-specified aperture and focus changes. Unlike paired benchmark evaluation, this setting does not have ground-truth target images, since the requested focus point and aperture value may correspond to a capture configuration that was never observed. We therefore provide qualitative examples to illustrate whether the model can respond to arbitrary user controls from a single input image.
Fig.~\ref{fig:demo} shows two types of edits. In the aperture-control examples, the focus point is kept fixed while the target aperture is varied. AnyBokeh consistently weakens or strengthens the defocus effect according to the specified aperture, even with existing bokeh in the input. In the focus-control examples, the aperture is kept fixed while the target focus point is moved to different scene locations. The model correspondingly shifts the sharp region to the selected focus point and adjusts foreground/background blur in a spatially coherent manner.

These examples show that AnyBokeh is not limited to reconstructing predefined source--target pairs. By combining the predicted source CoC and disparity with user-specified focus and aperture controls, our method can synthesize diverse post-capture bokeh edits from arbitrary inputs, supporting both aperture adjustment and focus relocation in a unified framework.

\subsection{Bokeh Rendering from AIF Inputs}

We further evaluate the conventional AIF-to-bokeh rendering setting to test whether AnyBokeh remains effective when the source blur signal is weak. For BokehMe and BokehDiff, we use the same global bokeh-level calibration protocol as in Sec.~\ref{sec:any2any}. For AnyBokeh, we run the full pipeline and compute the target CoC using relative optical fingerprint transfer. 
As shown in Tab.~\ref{tab:bokeh_rendering}, AnyBokeh achieves the best performance across all metrics on both datasets. This shows that, although AnyBokeh is designed for the more general any-to-any setting, it also performs strongly under the standard AIF-to-bokeh task. The results further indicate that the predicted source CoC and disparity provide sufficient guidance for relative target CoC computation even when the input is near all-in-focus. The second and fourth rows of Fig.~\ref{fig:any2any} showcase AIF-to-bokeh examples.

\begin{table}[t]
    \centering
    \footnotesize
    \caption{Bokeh rendering from all-in-focus inputs. 
    The same bokeh-level calibration protocol as the any-to-any setting is applied. AnyBokeh uses the full Stage~1+Stage~2 pipeline with relative target CoC transfer and requires no bokeh-level calibration. Best results are marked in \colorbox{cyan!20}{\textbf{bold}}.}
    \label{tab:bokeh_rendering}
    \setlength{\tabcolsep}{5.2pt}
    \renewcommand{\arraystretch}{1.0}
    \begin{tabular}{lccccccc}
    \toprule
    \multirow{2.5}{*}{Method} & \multirow{2}{*}{\makecell{No\\Calib.}} & \multicolumn{3}{c}{EBB!} & \multicolumn{3}{c}{RealBokeh} \\
    \cmidrule(lr){3-5}\cmidrule(lr){6-8}
    & & LPIPS$\downarrow$ & DISTS$\downarrow$ & FID$\downarrow$ & LPIPS$\downarrow$ & DISTS$\downarrow$ & FID$\downarrow$ \\
    \midrule
    BokehMe   & \xmark & 0.3682 & 0.1780 & 38.6188 & 0.3166 & 0.1439 & 59.8446 \\
    BokehDiff & \xmark & 0.4016 & 0.2411 & 58.0800 & 0.3309 & 0.1520 & 71.3716 \\
    \midrule
    \textbf{AnyBokeh (Ours)} & \cmark & \cellcolor{cyan!20}\textbf{0.3615} & \cellcolor{cyan!20}\textbf{0.1475} & \cellcolor{cyan!20}\textbf{36.0778} & \cellcolor{cyan!20}\textbf{0.3016} & \cellcolor{cyan!20}\textbf{0.1125} & \cellcolor{cyan!20}\textbf{45.8692} \\
    \bottomrule
    \end{tabular}
\end{table}

\begin{table}[t]
    \centering
    \footnotesize
    \caption{Defocus deblurring (bokeh-to-AIF) results. 
    No calibration is needed for all methods in this setting.
    Best results are marked in \colorbox{cyan!20}{\textbf{bold}}.}
    \label{tab:defocus_deblurring}
    \setlength{\tabcolsep}{6pt}
    \renewcommand{\arraystretch}{1.0}
    \begin{tabular}{lcccccc}
    \toprule
    \multirow{2.5}{*}{Method} & \multicolumn{3}{c}{EBB!} & \multicolumn{3}{c}{RealBokeh} \\
    \cmidrule(lr){2-4}\cmidrule(lr){5-7}
    & LPIPS$\downarrow$ & DISTS$\downarrow$ & FID$\downarrow$
    & LPIPS$\downarrow$ & DISTS$\downarrow$ & FID$\downarrow$ \\
    \midrule
    Restormer & \cellcolor{cyan!20}\textbf{0.4308} & 0.1943 & 56.5597 & \cellcolor{cyan!20}\textbf{0.3125} & 0.1352 & 57.1709 \\
    DRBNet    & 0.4320 & 0.1775 & 50.4203 & 0.3237 & 0.1302 & 51.3630 \\
    \midrule
    \textbf{AnyBokeh (Ours)} & 0.4406 & \cellcolor{cyan!20}\textbf{0.1378} & \cellcolor{cyan!20}\textbf{38.5373} & 0.3423 & \cellcolor{cyan!20}\textbf{0.1033} & \cellcolor{cyan!20}\textbf{44.1042} \\
    \bottomrule
    \end{tabular}
\end{table}

\begin{figure}[h]
    \centering
    \includegraphics[width=0.9\linewidth]{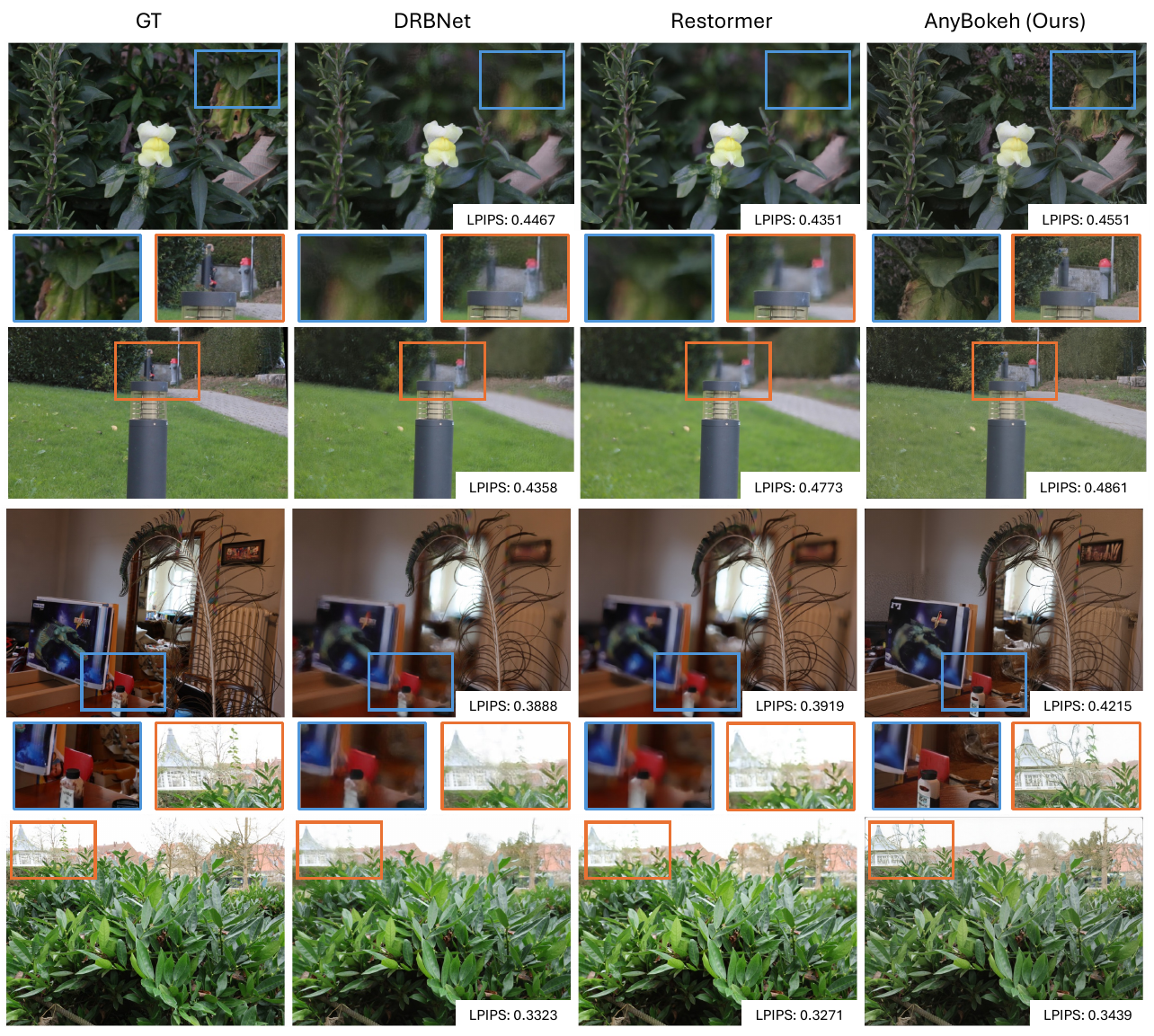}
    \caption{Qualitative examples illustrating the gap between LPIPS~\cite{lpips} and visual quality in defocus deblurring. Although AnyBokeh may obtain higher LPIPS than baselines, it often recovers sharper structures and more realistic local details in severely defocused regions. Restoration baselines can achieve lower LPIPS by producing smoother outputs, but they tend to preserve residual blur and under-recover fine structures.}
\label{fig:defocus_lpips}
    \label{fig:lpips}
\end{figure}

\subsection{Defocus Deblurring}
\label{sec:defocus_deblurring}

We evaluate defocus deblurring to examine whether AnyBokeh can effectively recover sharp structures when the target blur state requires removing existing defocus. We compare with Restormer~\cite{Zamir2021Restormer} and DRBNet~\cite{ruan2022drbnet}. As shown in Tab.~\ref{tab:defocus_deblurring}, AnyBokeh achieves the best DISTS and FID on both EBB! and RealBokeh, indicating improved perceptual structure and distribution-level realism.

AnyBokeh obtains slightly worse LPIPS than the restoration baselines. This discrepancy reflects a limitation of full-reference metrics for highly ill-posed generative restoration tasks. When an input region is severely defocused, the original high-frequency details are missing. A generative editor may recover sharper structures and synthesize visually plausible textures, but these details do not necessarily match the exact reference image at the pixel or feature level. As a result, LPIPS can penalize perceptually valid high-frequency recovery if the synthesized details differ from the ground truth.
Fig.~\ref{fig:lpips} illustrates this behavior. Compared with Restormer~\cite{Zamir2021Restormer} and DRBNet~\cite{ruan2022drbnet}, AnyBokeh often restores clearer object boundaries and more detailed local structures, especially in heavily blurred regions highlighted by the zoomed crops. However, because the recovered details are not guaranteed to be identical to the reference, the LPIPS score can be worse even when the visual result appears sharper and more realistic. In contrast, the restoration baselines tend to produce smoother outputs with residual defocus blur. Such outputs may remain closer to the reference in the LPIPS feature space, but they often under-recover fine structures and appear less visually sharp.
Therefore, LPIPS should be interpreted with caution in this setting. It remains useful for measuring reference similarity, but it does not fully capture the perceptual quality of plausible detail recovery under severe defocus.

\subsection{Ablation Studies}
\label{sec:ablation}

\begin{table}[t]
    \centering
    \footnotesize
    \caption{Ablation studies. We evaluate the effect of the two-stage design and source CoC conditioning. For both ablations, no tiling is used. All inputs are resized by the short edge and center-cropped. Best results are marked in \colorbox{cyan!20}{\textbf{bold}}.}
    \label{tab:ablation}
    \setlength{\tabcolsep}{6pt}
    \renewcommand{\arraystretch}{1.0}
    \begin{tabular}{llcccc}
    \toprule
    Ablation & Method & No Calib. & LPIPS$\downarrow$ & DISTS$\downarrow$ & FID$\downarrow$ \\
    \midrule
    \multirow{2}{*}{Pipeline}
    & Single-stage & \xmark & \cellcolor{cyan!20}\textbf{0.3058} & 0.1425 & 77.8112 \\
    & \textbf{Two-stage (Ours)} & \cmark & 0.3110 & \cellcolor{cyan!20}\textbf{0.1363} & \cellcolor{cyan!20}\textbf{76.8071} \\
    \midrule
    \multirow{2}{*}{Source CoC}
    & w/o Source CoC & \cmark & 0.3118 & 0.1413 & 81.0614 \\
    & \textbf{w. Source CoC (Ours)} & \cmark & \cellcolor{cyan!20}\textbf{0.3110} & \cellcolor{cyan!20}\textbf{0.1363} & \cellcolor{cyan!20}\textbf{76.8071} \\
    \bottomrule
    \end{tabular}
\end{table}

\begin{figure}
    \centering
    \includegraphics[width=0.8\linewidth]{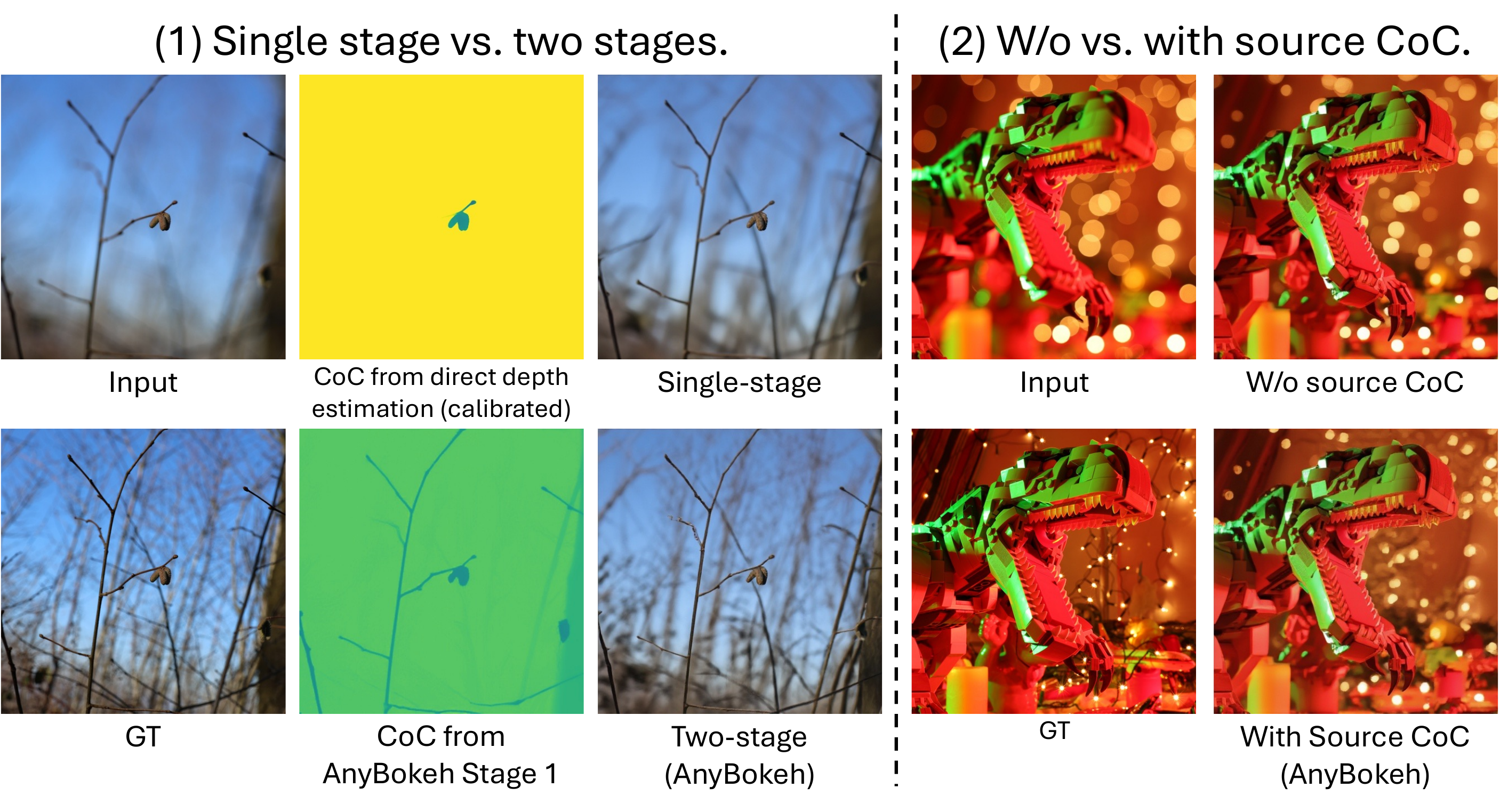}
    \caption{Visual comparisons for ablation studies. We evaluate the effect of the two-stage design and source CoC conditioning. For both ablations, no tiling is used. All inputs are resized by the short edge and center-cropped.}
    \label{fig:ablation}
\end{figure}

\textbf{One-stage vs. Two-stage.}
We ablate the necessity of our two-stage design by comparing it against a single-stage alternative that follows a more conventional depth-based pipeline. This variant first estimates monocular depth from the source image, then computes both source and target CoC maps from the estimated depth and EXIF metadata, and finally feeds these maps together with the source image into Stage~2. For target CoC computation, we use the source depth with the target optical settings, since the target image is unavailable during inference, and using target depth would introduce exposure bias. We use Depth-Pro~\cite{depthpro} for high-quality monocular depth estimation. We evaluate this variant on 100 challenging RealBokeh pairs with strong source defocus, where the source f-number $<f/4.0$. Because the single-stage variant does not recover a source-specific optical fingerprint from the observed blur--disparity relation, we apply the same global bokeh-level calibration protocol as in Sec.~\ref{sec:any2any}.
As shown in Tab.~\ref{tab:ablation}, the two-stage design improves DISTS and FID over the single-stage baseline, indicating better perceptual structure and realism. The single-stage variant obtains slightly better LPIPS, likely due to the same reason discussed in Sec.~\ref{sec:defocus_deblurring}. The single-stage variant relies on depth estimated directly from severely defocused inputs, leading to inaccurate scene geometry understanding and suboptimal editing. In contrast, our Stage~1 jointly predicts $CoC_{src}$ and disparity from the defocused input, allowing the model to learn blur-aware geometry and estimate a transferable optical fingerprint. See Fig.~\ref{fig:ablation}~(1) for visual comparison.

\textbf{With vs. Without Source CoC.}
We further ablate dual-CoC conditioning in Stage~2 by removing $CoC_{src}$ from the input while keeping $CoC_{tgt}$ unchanged. This variant tests whether the target blur specification alone is sufficient for bokeh synthesis.
As shown in Tab.~\ref{tab:ablation}, using $CoC_{src}$ improves all metrics. Without $CoC_{src}$, the editor must infer the input blur state only from image appearance, which is ambiguous in regions with spatially varying or severe defocus. Providing both $CoC_{src}$ and $CoC_{tgt}$ explicitly specifies the blur-state transition, giving the model stronger guidance for more accurate editing, as shown in Fig.~\ref{fig:ablation}~(2).
\section{Discussion and Limitations}
\label{app:limitation}

\begin{figure}
    \centering
    \includegraphics[width=0.9\linewidth]{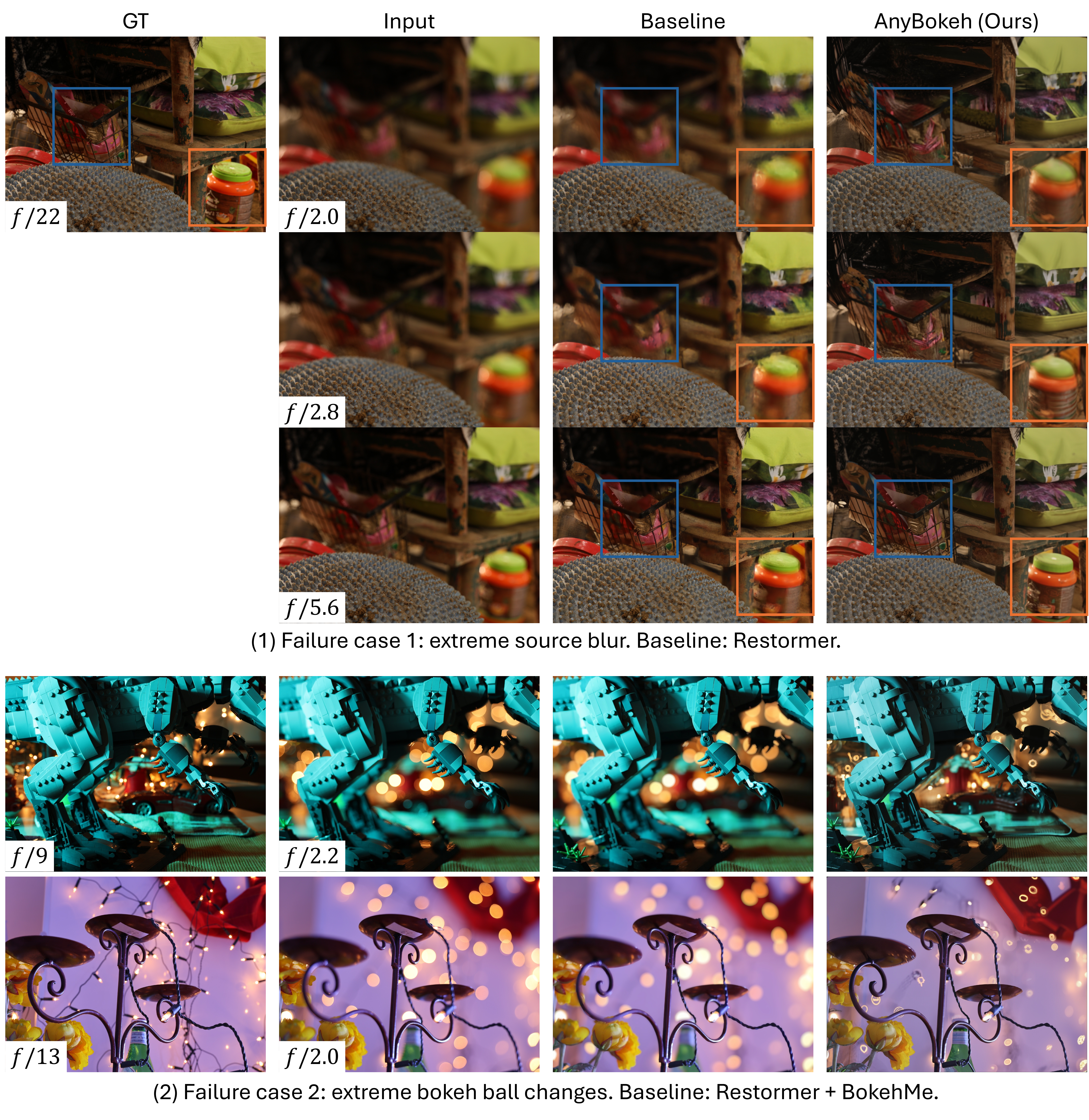}
    \caption{Failure cases under extreme source blur. (1) When the input is severely defocused (first row), missing high-frequency details limit a fully sharp reconstruction, although AnyBokeh improves as the source blur decreases (second and third rows) and remains clearer than the baseline. (2) When very large bokeh balls must be substantially reduced, AnyBokeh may leave residual or hollow bokeh artifacts, while the baseline often fails under such severe defocus. For both cases, the baseline with best quantitative results is used for comparison.}
    \label{fig:failure}
\end{figure}

AnyBokeh is built on the observation that the blur already present in a source image is not merely a degradation to be removed, but also carries useful optical information. By estimating the source CoC and disparity, our method transfers this source-specific optical fingerprint to the desired target setting and performs editing relative to the input blur state. This formulation avoids the need to canonicalize arbitrary inputs into an all-in-focus image, and our experiments show that explicitly conditioning on both source and target CoC provides effective guidance for spatially adaptive deblurring, preservation, and defocus synthesis.

Despite these advantages, AnyBokeh still has limitations.
First, the method is bound by the amount of visual evidence preserved in the source image. When the input is extremely defocused, high-frequency textures and fine geometric structures are missing, making faithful reconstruction inherently ambiguous. As shown in Fig.~\ref{fig:failure}~(1), AnyBokeh may not fully recover sharp details from severely blurred inputs, although the reconstruction improves as the source blur becomes weaker and remains clearer than the baseline. 
A related failure mode occurs for extremely large bokeh balls, especially when the target setting requires a much larger f-number and therefore a substantial reduction of the bokeh size. In such cases, shown in Fig.~\ref{fig:failure}~(2), the model may leave residual bokeh artifacts or produce hollow structures. This is partly because such extreme bokeh balls are rare in our training data, and partly because the underlying scene content is heavily occluded by the large defocus disk. The limited coverage of such extreme bokeh balls in our training data can also affect the reverse direction: when the target setting requires enlarging small highlights into very large bokeh balls, the generated bokeh size may be slightly underestimated. However, this case usually results in less pronounced bokeh rather than structural artifacts. Nevertheless, AnyBokeh remains more stable than the deblur-then-render baseline, which often fails under these severe blur conditions.
Second, our high-resolution inference relies on a tiling-and-stitching strategy to preserve the native aspect ratio and resolution of arbitrary input images. While this improves output quality and avoids resizing the whole image to a fixed resolution, it increases runtime compared with single-pass inference. Future work could improve efficiency with more compact backbones, better tile fusion strategies, or native high-resolution generative architectures. 
 
Overall, these limitations arise mainly in extreme blur or high-resolution settings, while AnyBokeh remains effective for a wide range of practical post-capture bokeh editing scenarios.
\section{Conclusion}
\label{sec:conclusion}

We introduced \textbf{AnyBokeh}, a physics-guided framework for any-to-any bokeh editing. Rather than forcing arbitrary inputs through an all-in-focus intermediate, AnyBokeh treats the source blur state as useful optical context and transfers its CoC-based optical fingerprint to the desired target focus and aperture setting. With dual-CoC conditioning, the editor explicitly models the relative transition between source and target blur states, enabling spatially adaptive bokeh editing without per-image bokeh-level calibration.
Experiments across any-to-any editing, AIF-to-bokeh rendering, and bokeh-to-AIF deblurring demonstrate the effectiveness of this formulation. Our ablations further show that both source CoC estimation and dual-CoC conditioning are important for faithful relative bokeh editing. 
Together with the proposed \textbf{UnrealBokeh} dataset, which provides per-pixel depth and complete EXIF metadata for computing ground-truth signed CoC maps, our work offers a practical step toward physically grounded post-capture DoF editing from in-the-wild images.

\newpage
{
    \small
    \bibliographystyle{ieeenat_fullname}
    \bibliography{main}

\begin{thebibliography}{44}
\providecommand{\natexlab}[1]{#1}
\providecommand{\url}[1]{\texttt{#1}}
\expandafter\ifx\csname urlstyle\endcsname\relax
  \providecommand{\doi}[1]{doi: #1}\else
  \providecommand{\doi}{doi: \begingroup \urlstyle{rm}\Url}\fi

\bibitem[Abuolaim and Brown(2020)]{abuolaim2020defocus}
Abdullah Abuolaim and Michael~S Brown.
\newblock Defocus deblurring using dual-pixel data.
\newblock In \emph{European conference on computer vision (ECCV)}, 2020.

\bibitem[Bando and Nishita(2007)]{bando2007towards}
Yosuke Bando and Tomoyuki Nishita.
\newblock Towards digital refocusing from a single photograph.
\newblock In \emph{Pacific Conference on Computer Graphics and Applications}, pages 363--372, 2007.

\bibitem[Barron et~al.(2015)Barron, Adams, Shih, and Hernández]{barron2015fast}
Jonathan~T. Barron, Andrew Adams, YiChang Shih, and Carlos Hernández.
\newblock Fast bilateral-space stereo for synthetic defocus.
\newblock In \emph{Proceedings of the IEEE/CVF Conference on Computer Vision and Pattern Recognition (CVPR)}, 2015.

\bibitem[Bochkovskii et~al.(2025)Bochkovskii, Delaunoy, Germain, Santos, Zhou, Richter, and Koltun]{depthpro}
Aleksei Bochkovskii, Ama\"{e}l Delaunoy, Hugo Germain, Marcel Santos, Yichao Zhou, Stephan~R. Richter, and Vladlen Koltun.
\newblock {Depth Pro}: Sharp monocular metric depth in less than a second.
\newblock In \emph{International Conference on Learning Representations (ICLR)}, 2025.

\bibitem[Busam et~al.(2019)Busam, Hog, McDonagh, and Slabaugh]{busam2019sterefo}
Benjamin Busam, Matthieu Hog, Steven McDonagh, and Gregory Slabaugh.
\newblock {SteReFo}: Efficient image refocusing with stereo vision.
\newblock In \emph{Proceedings of the IEEE/CVF International Conference on Computer Vision Workshops (ICCVW)}, 2019.

\bibitem[Chen et~al.(2025)Chen, Xie, Peng, Sun, Su, Yang, and Liu]{chen2025quadpixel}
Hang Chen, Yin Xie, Xiaoxiu Peng, Lihu Sun, Wenkai Su, Xiaodong Yang, and Chengming Liu.
\newblock Quad-pixel image defocus deblurring: A new benchmark and model.
\newblock In \emph{Proceedings of the IEEE/CVF Conference on Computer Vision and Pattern Recognition (CVPR)}, 2025.

\bibitem[Ding et~al.(2022)Ding, Ma, Wang, and Simoncelli]{dists}
Keyan Ding, Kede Ma, Shiqi Wang, and Eero~P. Simoncelli.
\newblock Image quality assessment: Unifying structure and texture similarity.
\newblock \emph{IEEE Transactions on Pattern Analysis and Machine Intelligence (PAMI)}, 44\penalty0 (5):\penalty0 2567--2581, 2022.

\bibitem[{Epic Games}()]{unrealengine}
{Epic Games}.
\newblock Unreal engine.

\bibitem[Fortes et~al.(2025)Fortes, Wei, Zhou, and Pan]{fortes2025bokehdiffusion}
Armando Fortes, Tianyi Wei, Shangchen Zhou, and Xingang Pan.
\newblock {Bokeh Diffusion}: Defocus blur control in text-to-image diffusion models.
\newblock In \emph{Proceedings of the SIGGRAPH Asia 2025 Conference Papers}, 2025.

\bibitem[Hach et~al.(2015)Hach, Steurer, Amruth, and Pappenheim]{hach2015cinematic}
Thomas Hach, Johannes Steurer, Arvind Amruth, and Artur Pappenheim.
\newblock Cinematic bokeh rendering for real scenes.
\newblock In \emph{Proceedings of the 12th European Conference on Visual Media Production}, 2015.

\bibitem[Haeberli and Akeley(1990)]{haeberli1990accumulation}
Paul Haeberli and Kurt Akeley.
\newblock The accumulation buffer: hardware support for high-quality rendering.
\newblock \emph{International Conference on Computer Graphics and Interactive Techniques}, 1990.

\bibitem[Heusel et~al.(2017)Heusel, Ramsauer, Unterthiner, Nessler, and Hochreiter]{fid}
Martin Heusel, Hubert Ramsauer, Thomas Unterthiner, Bernhard Nessler, and Sepp Hochreiter.
\newblock Gans trained by a two time-scale update rule converge to a local nash equilibrium.
\newblock In \emph{Advances in Neural Information Processing Systems}, page 6629–6640, 2017.

\bibitem[Hu et~al.(2022)Hu, Shen, Wallis, Allen-Zhu, Li, Wang, Wang, and Chen]{hu2022lora}
Edward~J Hu, Yelong Shen, Phillip Wallis, Zeyuan Allen-Zhu, Yuanzhi Li, Shean Wang, Lu Wang, and Weizhu Chen.
\newblock Lo{RA}: Low-rank adaptation of large language models.
\newblock In \emph{International Conference on Learning Representations (ICLR)}, 2022.

\bibitem[Ignatov et~al.(2020)Ignatov, Patel, and Timofte]{ignatov2020rendering}
Andrey Ignatov, Jagruti Patel, and Radu Timofte.
\newblock Rendering natural camera bokeh effect with deep learning.
\newblock In \emph{Proceedings of the IEEE Conference on Computer Vision and Pattern Recognition (CVPR) Workshops}, 2020.

\bibitem[Kraus and Strengert(2007)]{kraus2007depth}
M. Kraus and M. Strengert.
\newblock Depth-of-field rendering by pyramidal image processing.
\newblock \emph{Computer Graphics Forum}, 26\penalty0 (3):\penalty0 645--654, 2007.

\bibitem[Labs(2024)]{flux2024}
Black~Forest Labs.
\newblock Flux.
\newblock \url{https://github.com/black-forest-labs/flux}, 2024.

\bibitem[Lee et~al.(2021)Lee, Son, Rim, Cho, and Lee]{Lee2021IFAN}
Junyong Lee, Hyeongseok Son, Jaesung Rim, Sunghyun Cho, and Seungyong Lee.
\newblock Iterative filter adaptive network for single image defocus deblurring.
\newblock In \emph{Proceedings of the IEEE Conference on Computer Vision and Pattern Recognition (CVPR)}, 2021.

\bibitem[Lee et~al.(2009)Lee, Eisemann, and Seidel]{lee2009depth}
Sungkil Lee, Elmar Eisemann, and Hans-Peter Seidel.
\newblock Depth-of-field rendering with multiview synthesis.
\newblock \emph{ACM Transactions on Graphics}, 28\penalty0 (5):\penalty0 1–6, 2009.

\bibitem[Lee et~al.(2010)Lee, Eisemann, and Seidel]{lee2010realtime}
Sungkil Lee, Elmar Eisemann, and Hans-Peter Seidel.
\newblock Real-time lens blur effects and focus control.
\newblock \emph{ACM Transactions on Graphics}, 29:\penalty0 65:1--65:7, 2010.

\bibitem[Loshchilov and Hutter(2019)]{loshchilov2018adamw}
Ilya Loshchilov and Frank Hutter.
\newblock Decoupled weight decay regularization.
\newblock In \emph{International Conference on Learning Representations (ICLR)}, 2019.

\bibitem[Pan et~al.(2021)Pan, Chowdhury, Hartley, Liu, Zhang, and Li]{pan2021dual}
Liyuan Pan, Shah Chowdhury, Richard Hartley, Miaomiao Liu, Hongguang Zhang, and Hongdong Li.
\newblock Dual pixel exploration: Simultaneous depth estimation and image restoration.
\newblock In \emph{Proceedings of the IEEE/CVF Conference on Computer Vision and Pattern Recognition (CVPR)}, 2021.

\bibitem[Peebles and Xie(2023)]{Peebles2023DiT}
William Peebles and Saining Xie.
\newblock Scalable diffusion models with transformers.
\newblock In \emph{Proceedings of the IEEE/CVF International Conference on Computer Vision (ICCV)}, 2023.

\bibitem[Peng et~al.(2021)Peng, Luo, Xian, and Cao]{peng2021interactive}
Juewen Peng, Xianrui Luo, Ke Xian, and Zhiguo Cao.
\newblock Interactive portrait bokeh rendering system.
\newblock In \emph{IEEE International Conference on Image Processing (ICIP)}, 2021.

\bibitem[Peng et~al.(2022{\natexlab{a}})Peng, Cao, Luo, Lu, Xian, and Zhang]{Peng2022BokehMe}
Juewen Peng, Zhiguo Cao, Xianrui Luo, Hao Lu, Ke Xian, and Jianming Zhang.
\newblock {BokehMe}: When neural rendering meets classical rendering.
\newblock In \emph{Proceedings of the IEEE/CVF International Conference on Computer Vision and Pattern Recognition (CVPR)}, 2022{\natexlab{a}}.

\bibitem[Peng et~al.(2022{\natexlab{b}})Peng, Zhang, Luo, Lu, Xian, and Cao]{Peng2022MPIB}
Juewen Peng, Jianming Zhang, Xianrui Luo, Hao Lu, Ke Xian, and Zhiguo Cao.
\newblock {MPIB}: An mpi-based bokeh rendering framework for realistic partial occlusion effects.
\newblock In \emph{European conference on computer vision (ECCV)}, 2022{\natexlab{b}}.

\bibitem[Peng et~al.(2025)Peng, Cao, Luo, Xian, Tang, Zhang, and Lin]{bokehme++}
Juewen Peng, Zhiguo Cao, Xianrui Luo, Ke Xian, Wenfeng Tang, Jianming Zhang, and Guosheng Lin.
\newblock {BokehMe++}: Harmonious fusion of classical and neural rendering for versatile bokeh creation.
\newblock \emph{IEEE Transactions on Pattern Analysis and Machine Intelligence (PAMI)}, 47\penalty0 (3):\penalty0 1530--1547, 2025.

\bibitem[Potmesil and Chakravarty(1981)]{potmesil1981lens}
Michael Potmesil and Indranil Chakravarty.
\newblock A lens and aperture camera model for synthetic image generation.
\newblock In \emph{International Conference on Computer Graphics and Interactive Techniques}, 1981.

\bibitem[Quan et~al.(2023)Quan, Yao, and Ji]{quan2023single}
Yuhui Quan, Xin Yao, and Hui Ji.
\newblock Single image defocus deblurring via implicit neural inverse kernels.
\newblock In \emph{Proceedings of the IEEE/CVF International Conference on Computer Vision (ICCV)}, 2023.

\bibitem[Ruan et~al.(2022)Ruan, Chen, Li, and Lam]{ruan2022drbnet}
Lingyan Ruan, Bin Chen, Jizhou Li, and Miuling Lam.
\newblock Learning to deblur using light field generated and real defocus images.
\newblock In \emph{Proceedings of the IEEE/CVF conference on computer vision and pattern recognition (CVPR)}, 2022.

\bibitem[Seizinger et~al.(2025)Seizinger, Vasluianu, Conde, Wu, and Timofte]{seizinger2025bokehlicious}
Tim Seizinger, Florin-Alexandru Vasluianu, Marcos Conde, Zongwei Wu, and Radu Timofte.
\newblock Bokehlicious: Photorealistic bokeh rendering with controllable apertures.
\newblock In \emph{Proceedings of the IEEE/CVF International Conference on Computer Vision (ICCV)}, 2025.

\bibitem[Sheng et~al.(2024)Sheng, Yu, Ling, Cao, Zhang, Lu, Xian, Lin, and Benes]{sheng2024drbokeh}
Yichen Sheng, Zixun Yu, Lu Ling, Zhiwen Cao, Xuaner Zhang, Xin Lu, Ke Xian, Haiting Lin, and Bedrich Benes.
\newblock {Dr. Bokeh}: Differentiable occlusion-aware bokeh rendering.
\newblock In \emph{Proceedings of the IEEE/CVF Conference on Computer Vision and Pattern Recognition (CVPR)}, 2024.

\bibitem[Tang et~al.(2024)Tang, Xu, Zhou, Wei, Han, Cao, and Lasser]{tang2024prior}
Peng Tang, Zhiqiang Xu, Chunlai Zhou, Pengfei Wei, Peng Han, Xin Cao, and Tobias Lasser.
\newblock Prior and prediction inverse kernel transformer for single image defocus deblurring.
\newblock In \emph{Proceedings of the AAAI Conference on Artificial Intelligence}, 2024.

\bibitem[Tuan~Mu et~al.(2025)Tuan~Mu, Huang, and Liu]{Genfocus2025}
Chun-Wei Tuan~Mu, Jia-Bin Huang, and Yu-Lun Liu.
\newblock Generative refocusing: Flexible defocus control from a single image.
\newblock \emph{arXiv preprint arXiv:2512.16923}, 2025.

\bibitem[Wadhwa et~al.(2018)Wadhwa, Garg, Jacobs, Feldman, Kanazawa, Carroll, Movshovitz-Attias, Barron, Pritch, and Levoy]{scattering}
Neal Wadhwa, Rahul Garg, David~E. Jacobs, Bryan~E. Feldman, Nori Kanazawa, Robert Carroll, Yair Movshovitz-Attias, Jonathan~T. Barron, Yael Pritch, and Marc Levoy.
\newblock Synthetic depth-of-field with a single-camera mobile phone.
\newblock \emph{ACM Transactions on Graphics}, 37\penalty0 (4):\penalty0 1–13, 2018.

\bibitem[Wang et~al.(2025)Wang, Chen, Xu, Liu, and Zhao]{wang2025diffcamera}
Yiyang Wang, Xi Chen, Xiaogang Xu, Yu Liu, and Hengshuang Zhao.
\newblock {DiffCamera}: Arbitrary refocusing on images.
\newblock In \emph{Proceedings of the SIGGRAPH Asia 2025 Conference Papers}, 2025.

\bibitem[Wu et~al.(2012)Wu, Zheng, Hu, and Xu]{wu2012rendering}
Jiaze Wu, Changwen Zheng, Xiaohui Hu, and Fanjiang Xu.
\newblock Rendering realistic spectral bokeh due to lens stops and aberrations.
\newblock \emph{The Visual Computer}, 29, 2012.

\bibitem[Yang et~al.(2023)Yang, Pan, Liu, and Liu]{yang2023k3dn}
Yan Yang, Liyuan Pan, Liu Liu, and Miaomiao Liu.
\newblock {K3DN}: Disparity-aware kernel estimation for dual-pixel defocus deblurring.
\newblock In \emph{Proceedings of the IEEE/CVF Conference on Computer Vision and Pattern Recognition (CVPR)}, 2023.

\bibitem[Yu et~al.(2010)Yu, Wang, and Yu]{yu2010realtime}
Xuan Yu, Rui Wang, and Jingyi Yu.
\newblock Real-time depth of field rendering via dynamic light field generation and filtering.
\newblock \emph{Computer Graphics Forum}, 29\penalty0 (7):\penalty0 2099--2107, 2010.

\bibitem[Yuan et~al.(2024)Yuan, Wang, Sheng, Chennuri, Zhang, and Chan]{Yuan_2024_GenPhoto}
Yu Yuan, Xijun Wang, Yichen Sheng, Prateek Chennuri, Xingguang Zhang, and Stanley Chan.
\newblock Generative photography: Scene-consistent camera control for realistic text-to-image synthesis.
\newblock \emph{arXiv preprint arXiv: 2412.02168}, 2024.

\bibitem[Zamir et~al.(2022)Zamir, Arora, Khan, Hayat, Khan, and Yang]{Zamir2021Restormer}
Syed~Waqas Zamir, Aditya Arora, Salman Khan, Munawar Hayat, Fahad~Shahbaz Khan, and Ming-Hsuan Yang.
\newblock Restormer: Efficient transformer for high-resolution image restoration.
\newblock In \emph{Proceedings of the IEEE/CVF conference on computer vision and pattern recognition (CVPR)}, 2022.

\bibitem[Zhang et~al.(2018)Zhang, Isola, Efros, Shechtman, and Wang]{lpips}
Richard Zhang, Phillip Isola, Alexei~A. Efros, Eli Shechtman, and Oliver Wang.
\newblock The unreasonable effectiveness of deep features as a perceptual metric.
\newblock In \emph{Proceedings of the IEEE/CVF Conference on Computer Vision and Pattern Recognition (CVPR)}, 2018.

\bibitem[Zheng et~al.(2024)Zheng, Gao, Fan, Liu, Laaksonen, Ouyang, and Sebe]{zheng2024birefnet}
Peng Zheng, Dehong Gao, Deng-Ping Fan, Li Liu, Jorma Laaksonen, Wanli Ouyang, and Nicu Sebe.
\newblock Bilateral reference for high-resolution dichotomous image segmentation.
\newblock \emph{CAAI Artificial Intelligence Research}, 3:\penalty0 9150038, 2024.

\bibitem[Zhu et~al.(2025)Zhu, Fan, Zhang, Chen, Zhang, Xu, and Shi]{zhu2025bokehdiff}
Chengxuan Zhu, Qingnan Fan, Qi Zhang, Jinwei Chen, Huaqi Zhang, Chao Xu, and Boxin Shi.
\newblock {BokehDiff}: Neural lens blur with one-step diffusion.
\newblock In \emph{Proceedings of the IEEE/CVF International Conference on Computer Vision (ICCV)}, 2025.

\bibitem[Zhu et~al.(2013)Zhu, Cohen, Schiller, and Milanfar]{zhu2013estimating}
Xiang Zhu, Scott Cohen, Stephen Schiller, and Peyman Milanfar.
\newblock Estimating spatially varying defocus blur from a single image.
\newblock \emph{IEEE Transactions on Image Processing (TIP)}, 22\penalty0 (12):\penalty0 4879--4891, 2013.

\end{thebibliography}
}

\end{document}